\pdfoutput=1

\documentclass[11pt]{article}

\usepackage[preprint]{acl}

\usepackage{times}
\usepackage{latexsym}

\usepackage[T1]{fontenc}

\usepackage[utf8]{inputenc}

\usepackage{microtype}

\usepackage{inconsolata}

\usepackage{graphicx}

\usepackage{amsmath}
\usepackage{colortbl}
\usepackage{xcolor}
\usepackage{ulem}
\usepackage{rotating}
\usepackage{color}
\usepackage{subfigure}
\usepackage{multirow}
\usepackage{float}
\usepackage{amssymb}
\usepackage{booktabs}
\usepackage{algorithm}
\usepackage{algorithmic}
\usepackage{arydshln}
\usepackage{url}
\usepackage{subcaption}
\usepackage{graphicx}

\title{RefineCoder: Iterative Improving Large Language Models through Adaptive Critique Refinement for Code Generation}

\author{Changzhi Zhou$^{1}$\thanks{~ Work done during internship at Meituan. Equal contribution.}, \  Xinyu Zhang$^{1}$\footnotemark[1], \  Dandan Song$^{1}$\thanks{~ Corresponding Author.}, \ Xiancai Chen$^{2}$, Wanli Gu$^{3}$, \\ {\bf Huipeng Ma$^{1}$, Yuhang Tian$^{1}$, Mengdi Zhang$^{3}$, Linmei Hu$^{1}$\footnotemark[2]} \\
  $^{1}$School of Computer Science and Technology, \\Beijing Institute of Technology, Beijing, China \\ $^{2}$Peking University  \ \ $^{3}$Meituan \\
  \texttt{zhouchangzhi97@gmail.com}  \ \ \texttt{\{sdd,hulinmei\}@bit.edu.cn}}

\begin{document}
\maketitle
\begin{abstract}
Code generation has attracted increasing attention with the rise of Large Language Models (LLMs). Many studies have developed powerful code LLMs by synthesizing code-related instruction data and applying Supervised Fine-Tuning (SFT). However, these methods are limited by teacher model distillation and ignore the potential of iterative refinement by self-generated code. In this paper, we propose Adaptive Critique Refinement (ACR), which enables the model to refine itself by self-generated code response and teacher-generated critique, rather than directly imitating the teacher-generated code responses. Concretely, ACR includes a composite scoring system with LLM-as-a-Judge to evaluate the quality of code responses and a selective critiquing strategy with LLM-as-a-Critic to critique self-generated low-quality code responses. We develop the RefineCoder series by iteratively applying ACR, achieving continuous performance improvement on multiple code generation benchmarks. Compared to the baselines of the same size, our proposed RefineCoder series can achieve superior performance using less data.
\end{abstract}

\section{Introduction}

Code generation, also called program synthesis, is a key application area of Large Language Models (LLMs) and has attracted significant attention from the research community~\cite{gulwani2017program,chen2021evaluating}. Numerous studies focus on developing code-specific pre-trained models, such as CodeLlama~\cite{roziere2023code}, DeepSeek-Coder~\cite{guo2024deepseek}, and Qwen-Coder~\cite{hui2024qwen2}. Despite their strong capabilities in code understanding, these LLMs typically require additional post-training to become effective and user-friendly for code-related tasks.

\begin{figure}[t]
    \centering
    \includegraphics[width=1.0\linewidth]{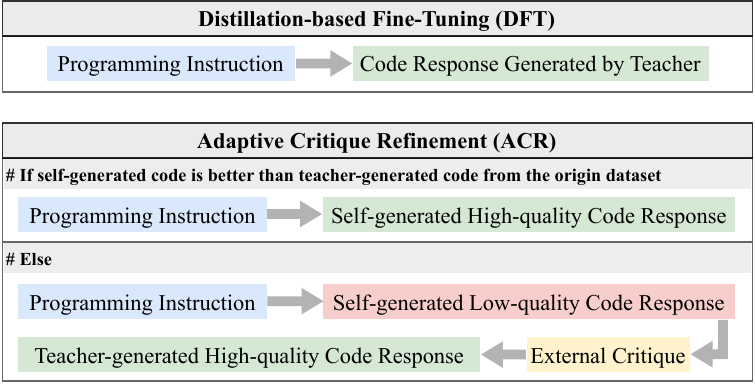}
    \caption{Comparison of two fine-tuning paradigms. DFT uses code distilled by teacher model, whereas ACR adaptively constructs distinct data formats infusing self-generated code and teacher-generated critique.}
    \label{intro:example}
\end{figure}

As a popular post-training technology, Distillation-based Fine-Tuning (DFT) leverages code instruction data synthesized by powerful teacher models (often proprietary LLMs like GPT-4) to fine-tune code LLMs, as shown in Figure~\ref{intro:example}. Many works~\cite{zhang2024unifyingpe} based on this paradigm have been proposed recently. For example, Code Alpaca~\cite{codealpaca} first establishes this synthesis paradigm using Self-Instruct~\cite{wang-etal-2023-self-instruct}. Subsequently, Code-related Evol-Instruct~\cite{luo2024wizardcoder} prompts LLMs to evolve more intricate code instructions, while OSS-Instruct~\cite{wei2024magicoder} generates more realistic code instructions based on open-source code snippets~\cite{kocetkov2022stack}. 
In contrast, OpenCodeInterpreter~\cite{zheng-etal-2024-opencodeinterpreter} enables code LLMs to learn from compiler diagnostics and human feedback by constructing multi-turn feedback data. These diverse synthesis strategies have significantly advanced the development of code LLMs. However, this paradigm of \textbf{\textit{teacher model distillation}} inevitably suffers from a critical limitation: the performance of the student model largely relies on the teacher model, ignoring the potential of iterative refinement by self-generated code.

Apart from research on code instruction fine-tuning, some other studies~\cite{chen2023codet,chen2024teaching,chen2025revisitselfdebuggingselfgeneratedtests, jiang2025logicproimprovingcomplexlogical} have explored the ability of LLMs to refine self-generated code during inference. For example, Self-Debugging~\cite{chen2024teaching} iteratively debugs self-generated code using feedback from the code executor. However, these methods that freeze parameters are essentially a form of \textbf{\textit{prompt engineering with external feedback}}. They cannot improve the intrinsic one-off code generation ability of LLMs. Besides,  multiple calls to LLMs increase inference latency.

In this paper, we propose a novel fine-tuning paradigm called Adaptive Critique Refinement (ACR). It evolves through a combination of self-generated code and teacher-driven critique, thereby enhancing its capacity for one-shot code generation and surpassing the limitations of conventional teacher distillation methods. Specifically, as shown in Figure~\ref{intro:example}, ACR evaluates both self-generated and teacher-generated code, adaptively critiques low-quality self-generated responses, and generates new samples based on the evaluation and critique results. This process is analogous to a student initially attempting to solve problems independently and later comparing their solutions against given answers, rather than simply copying the provided solutions. By iteratively applying the ACR process, the model achieves continuous improvement in its code generation capabilities. Notably, we design a composite scoring system and a selective critique strategy to perform scoring and critique, respectively. The scoring system combines LLM-as-a-Judge and a code executor to provide accurate and comprehensive evaluations. The critique strategy adaptively constructs two types of data and incorporates LLM-as-a-Critic.

We conduct iterative ACR based on DS-Coder-6.7B-Base~\cite{guo2024deepseek} and Qwen2.5-Coder-7B-Base~\cite{hui2024qwen2}, leading to the development of the RefineCoder series. After three iterations, RefineCoder-DS-6.7B and RefineCoder-QW-7B demonstrate significant improvements in average pass@1 performance on the HumanEval(+), MBPP(+), LiveCodeBench, and BigCodeBench (hard)~\cite{chen2021evaluating,austin2021programsynthesislargelanguage,liu2023is,jain2024livecodebench,zhuo2024bigcodebench}. The key contributions of this paper are as follows:

1) We propose Adaptive Critique Refinement (ACR), a novel fine-tuning paradigm that refines code LLMs using self-generated code and external critique, on the basis of which we develop a series of strong code LLMs namely RefineCoder.

2) To ensure the effectiveness of ACR, we design a composite scoring system with LLM-as-a-Judge and a selective critiquing strategy with LLM-as-a-Critic, enabling both accurate scoring and critique of code responses.

3) Experimental results from the RefineCoder series show that iterative ACR continuously improves code generation performance. After three iterations, the RefineCoder models outperform baselines of the same size while requiring less data.

\section{Related Work}

\textbf{LLMs for Code Generation} LLMs have shown exceptional code understanding abilities due to extensive pre-training on code-related corpora. Numerous models like GPT-4o~\cite{gpt4o-blog}, Gemini~\cite{gemini-blog}, Claude-3.5~\cite{claude3p5-blog}, Qwen2.5-Coder~\cite{hui2024qwen2}, and DeepSeek-Coder~\cite{guo2024deepseek} exhibit strong performance on code generation benchmarks. Recently, the release of OpenAI o1~\cite{o1-blog} and DeepSeek-R1~\cite{2025deepseekr1} has spurred a surge in research on deep reasoning LLMs, achieving expert-level performance on competitive programming problems (e.g., CodeForces) and further advancing LLMs in the code domain.

\begin{figure*}[t]
    \centering
    \includegraphics[width=2.0\columnwidth]{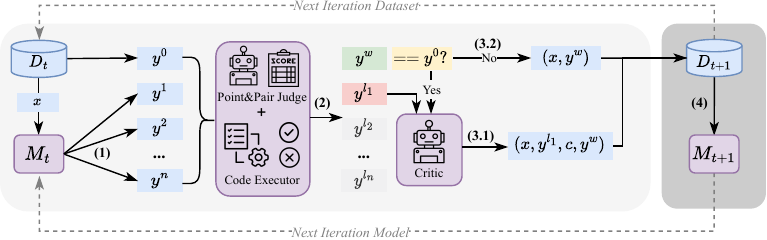}
    \caption{\textbf{Overview of ACR in the $t$-th iteration.} \textbf{(1) Sampling}: The model $M_t$ samples $n$ code responses $\{y^{i}\}_{i=1}^n$ with high temperature for a programming instruction $x$ in the dataset $D_t$. \textbf{(2) Ranking}: A composite scoring system first scores all self-generated code responses $\{y^{i}\}_{i=1}^n$, then selects the highest-scoring response for pairwise comparison with the original code response $y^0$ to identify the winner $y^w$ and the loser $y^{l_1}$. \textbf{(3) Refining}: Depending on the identity of the $y^w$, the selective critiquing strategy constructs a new single-turn data $(x,y^w)$ or two-turn critique data $(x, y^{l_1}, c, y^w)$, where $c$ is critique. The original data $(x, y)$ is replaced with new data and all the new data form a new dataset $D_{t+1}$. \textbf{(4) Training}: The new model $M_{t+1}$ is fine-tuned using $D_{t+1}$.}
	\label{method:model}
\end{figure*}

\textbf{Distillation-based Code Fine-Tuning} 
Unlike proprietary models, many studies focus on fine-tuning open-source code pre-trained models, which has greatly contributed to the rapid development of the code generation field. A key technique for achieving this is distilling data from teacher models. Code Alpaca~\cite{codealpaca} introduces Self-Instruct~\cite{wang-etal-2023-self-instruct} to distill GPT-3. Following this, WizardCoder~\cite{luo2024wizardcoder} evolves more complex code instruction data using Evol-Instruct~\cite{xu2024wizardlm}. MagiCoder~\cite{wei2024magicoder} proposes OSS-Instruct, where the model generates instructions and code responses sequentially based on open-source code snippets, thus creating more practical code data. In contrast to OSS-Instruct, InverseCoder~\cite{wu2024inversecoders} reverses the order of instruction and code response generation. WaveCoder~\cite{yu-etal-2024-wavecoder} constructs a multi-task code dataset, enhancing the model's versatility. Besides, OpenCodeInterpreter~\cite{zheng-etal-2024-opencodeinterpreter} builds multi-turn code data, enabling the model to learn from execution feedback and human feedback. However, these methods aim to enable the model to learn by imitating the teacher, overlooking the potential for refinement through self-generated code.

\textbf{Iterative Self-Refinement of LLMs} 
Iterative self-refinement refers to LLMs enhancing themselves iteratively by utilizing self-generated responses with the help of external signals. 
One line of research~\cite{huang-etal-2023-large,madaan2023selfrefine,hu-etal-2024-teaching} focuses on enabling self-correction during the inference stage by iteratively calling LLMs and incorporating external signals. In the code domain, CodeT~\cite{chen2023codet}, Self-Debugging~\cite{chen2024teaching}, and LDB~\cite{zhong-etal-2024-debug} follow this approach. However, these prompt engineering methods with external feedback cannot improve the intrinsic capabilities. Another line of research focuses on iteratively training the model using self-generated outputs to enhance its intrinsic capabilities~\cite{dong2024self,yuan2024selfrewarding,s3cmath,kim2025spread}. These methods typically rely on preference learning, such as DPO~\cite{rafailov2023direct}. CodeLutra~\cite{tao2024codelutra} has successfully applied this approach to the code domain, but it heavily depends on golden labels, which limits its applicability. In contrast to the works above, we propose the ACR method, which achieves iterative self-refinement by only using a simple Supervised Fine-Tuning (SFT) loss. Our approach is orthogonal to prompt engineering methods like Self-Debugging and is more efficient and generalizable than CodeLutra.

\section{Methodology}

\subsection{Overview}

Adaptive Critique Refinement (ACR) aims to improve code LLMs by refining the existing dataset with self-generated code and external critique. The iterative application of ACR facilitates continuous improvement in code generation performance. As illustrated in Figure~\ref{method:model}, ACR starts with a current dataset and a code model fine-tuned on it. The method updates this dataset through a process of sampling, ranking, and refining. The updated dataset is used for the next round of fine-tuning, iteratively improving the model's code generation capabilities. 

In the following sections, we will introduce two key components of the pipeline above: the composite scoring system for ranking and the selective critiquing strategy for data refining. Finally we describe the iterative training and the resulting RefineCoder model.



\subsection{Composite Scoring System with LLM-as-a-Judge}

A comprehensive and accurate evaluation of code response quality is the foundation for effective ACR. Previous works~\cite{chen2023codet,zhang2024codedpo} rely on generating test cases to evaluate code quality. However, this method has issues with inaccurate test case generation~\cite{jain2024livecodebench}, inability to handle \textit{special judge}~\cite{quan2025codeelo}, and failure to evaluate non-code content in response such as code comments and explanations.

Drawing on the judgment and evaluation abilities of LLMs~\cite{dong2024self,li2024llmasajudge}, we employ LLM-as-a-Judge as the backbone of the scoring system, supplemented by a code executor to ensure effective evaluation. Specifically, 
we perform a multi-faceted, point-wise judgment of each self-generated code response $y^i$ ($1 \le i \le n$):

\begin{equation}
s^i_1, s^i_2, ..., s^i_m = \textrm{Multi-Faceted Judge}(x, y^i),
\end{equation}

where $s^i_j$ denotes the score for $y^i$ under the $j$-th evaluation criterion, and $m$ is the total number of criteria (e.g., code correctness, clarity). Next, we calculate the final point-wise score for each self-generated code response by computing the weighted sum and normalizing the result:

\begin{align}
s^i &= \sum_{j=1}^{m} \alpha_j s_j^i, \\
\text{score}^i_{\text{point}} &= \frac{s^i - s_{\min}}{s_{\max} - s_{\min}},
\end{align}

where $\alpha_j$ is the weight of the $j$-th criterion, and $s_{\min}$ and $s_{\max}$ are the minimum and maximum scores across self-generated code responses. In addition, we use a code executor to evaluate the executability of all code responses (including $y^0$):

\begin{equation}
\textrm{score}_{\text{exec}}^i = \textrm{Executor}(y^i),
\end{equation}

where $i \in [0, n]$ and $\textrm{score}_{\text{exec}}^i \in \{0, 1\}$. The highest-scoring self-generated code is obtained by summing the point-wise score and execution score: 

\begin{equation}
y^{self} = \arg \max_i (\text{score}^i_{\text{point}} + \text{score}^i_{\text{exec}}).
\end{equation}

where $i \in [1,n]$. 
Finally, we determine the winner between the best self-generated code $y^{self}$ and the teacher-generated code $y^0$ based on the pairwise judgment and execution scores:

\begin{equation}
    \textrm{score}_{\text{pair}}^{self}, \textrm{score}_{\text{pair}}^{0} = \textrm{PairJudge}(x, y^{self}, y^0),
\end{equation}

\begin{equation}
    \textrm{score}^{i} = \textrm{score}^{i}_{\text{pair}} + \textrm{score}^{i}_{\text{exec}},
\end{equation}

\begin{equation}
    y^w, y^{l_1} = \begin{cases}
y^{self}, y^0, & \text{if } \textrm{score}^{self}  > \textrm{score}^{0}, \\
 y^0, y^{self}, & \text{if } \textrm{score}^{0}  > \textrm{score}^{self},
\end{cases}    
\end{equation}

where $i \in \{self, 0\}$, $\textrm{score}^i_{pair} \in [0, 1]$, and $y^w$ and $y^{l_1}$ denote the winner and loser, respectively. The prompts for point-wise and pair-wise judgment are shown in Figures~\ref{prompt:pointjudge} and~\ref{prompt:pairjudge}.

\subsection{Selective Critiquing Strategy with LLM-as-a-Critic}

After selecting winner and loser based on quality, we design the selective critiquing strategy as the data construction engine of ACR. It effectively utilizes high-quality and low-quality self-generated code in distinct ways, while further providing external critique for the low-quality data. This strategy mirrors the process of a student first solving problems and then critiquing their answers based on provided solutions rather than directly imitating the solutions~\cite{wang2025critique}. 

Concretely, when all self-generated code responses are of lower quality than the original code in the current dataset (i.e., $y^w$ is $y^0$ from origin dataset and $y^{l_1}$ is the best self-generated code response), LLM-as-a-Critic is used to critique $y^{l_1}$, using $y^0$ as a reference (The prompt is shown in Figure~\ref{prompt:paircritic}):

\begin{equation}
c = \textrm{Critic}(x, y^w, y^{l_1}),
\end{equation}

where $c$ is the critique of $y^{l_1}$. Otherwise, if $y^w$ is the self-generated code response, we directly construct new single-turn data. The data update rules are as follows:

\begin{equation}
(x, y) \longrightarrow \begin{cases}
(x, y^{l_1}, c, y^w), & \text{if } y^w \text{ is } y^0, \\
(x, y^w), & \text{otherwise}.
\end{cases}
\end{equation}

If the data to be updated comes from a two-turn data rather than a single-turn one, only the instruction and the final code response will be considered. The selective critiquing strategy updates all data to comprise a new instruction dataset $D_{t+1}$.

\subsection{Iterative Training}

The previous sections describe the process of a single ACR in Figure~\ref{method:model}. 
Next, we define the RefineCoder in the iterative ACR process:

$\circ$ \textbf{RefineCoder} \textit{Iter0} ($M_0$): The code model is trained using the initial SFT dataset $D_0$.

$\circ$ \textbf{RefineCoder}  \textit{Iter1} ($M_1$): The code model is trained using dataset $D_1$, where $D_1$ is obtained by refining $D_0$ using $M_0$.

$\circ$ \textbf{RefineCoder}  \textit{Itert} ($M_t$): The code model is trained using dataset $D_t$, where $D_t$ is obtained by refining $D_{t-1}$ using $M_{t-1}$.

We fine-tune the original pre-trained model from scratch in each iteration to prevent overfitting, consistent with previous iterative training works~\cite{wang2024selftaughtevaluators,dong2024self}.

\begin{table*}[t]
\centering
\begin{tabular}{lccccccll}
    \toprule[1.5pt]
    \multicolumn{1}{l}{\multirow{2}[2]{*}{\textbf{Models}}}  & \multicolumn{1}{l}{\multirow{2}[2]{*}{\textbf{Data Size}}} & \multicolumn{2}{c}{\textbf{LiveCodeBench}} & \multicolumn{2}{c}{\textbf{BigCodeBench-hard}} & \multirow{2}[2]{*}{\textbf{Average}} \\
\cline{3-4} \cline{5-6}  

& & \multicolumn{1}{c}{\textit{v3-v5}} & \multicolumn{1}{c}{\textit{v5-v6}} & \multicolumn{1}{c}{\textit{Instruct}} & \multicolumn{1}{c}{\textit{Complete}} & \\    
    \midrule[1.5pt]
\raisebox{-0.1cm}{\includegraphics[width=0.5cm]{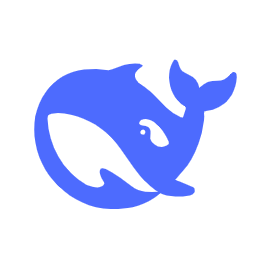}} DSCoder-6.7B-Base & / & / & / & / & 13.5 & /   \\
DSCoder-6.7B-Instruct & 2BT  &  12.8   & 14.0 & 10.1              & 15.5    & 13.1        \\
MagiCoder-S-DS-6.7B  & 185K  &13.4 & 14.6 & 13.5        &       12.8    & 13.6   \\
OpenCodeInterpreter-DS-6.7B & 178K &  7.3  &  15.2 & 13.5              & \textbf{16.9}  & 13.2     \\
WaveCoder-Ultra-6.7B    & 130K & 12.9 & 14.5 &  12.8              & \textbf{16.9}   &          14.3 \\
\rowcolor{gray!15} \textbf{RefineCoder-DS-6.7B} \ \ \textit{Iter0}       & 80K       &  13.1   &  15.5 & 18.2            & 10.1   & 14.2   \\
\rowcolor{gray!30}  \textbf{RefineCoder-DS-6.7B} \ \ \textit{Iter1}   & 80K             & \textbf{14.3} & 17.2           & 18.9              & 14.2         & 16.2\textcolor[rgb]{0,0.5,0}{$_{\uparrow2.0}$}                \\
\rowcolor{gray!45} \textbf{RefineCoder-DS-6.7B} \ \ \textit{Iter2}    & 80K          &  \textbf{14.3}  &        17.8           & 19.6              & 14.2           & 16.5\textcolor[rgb]{0,0.5,0}{$_{\uparrow2.3}$} \\
\rowcolor{gray!60} \textbf{RefineCoder-DS-6.7B} \ \ \textit{Iter3}   & 80K    & 14.2  & \textbf{17.9}  & \textbf{20.1}              & 15.5 & \textbf{16.9}\textcolor[rgb]{0,0.5,0}{$_{\uparrow2.7}$}  \\
\midrule[1pt]
\raisebox{-0.1cm}{\includegraphics[width=0.5cm]{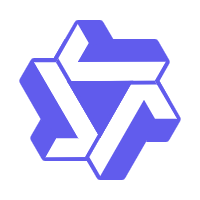}}Qwen2.5-Coder-7B-Base  & /  & /    & /    & /    & 16.2                                    & \ \ \ / \\
Qwen2.5-Coder-7B-Instruct & Millions & 18.1 & 18.9 & 20.3 & 20.3 & 19.4  \\
\rowcolor{gray!15} \textbf{RefineCoder-QW-7B} \ \ \textit{Iter0}       & 80K      &  18.0  &   18.7                    & 18.5            & 19.3   & 18.6   \\
\rowcolor{gray!30} \textbf{RefineCoder-QW-7B} \ \ \textit{Iter1}    & 80K   &   19.6 & 21.3                      & 20.7              & 21.8           & 20.9\textcolor[rgb]{0,0.5,0}{$_{ \uparrow2.3}$} \\
\rowcolor{gray!45} \textbf{RefineCoder-QW-7B} \ \ \textit{Iter2}   & 80K     &   20.5   &            21.5     & 20.6      & \textbf{22.9}      & 21.4\textcolor[rgb]{0,0.5,0}{$_{\uparrow2.8}$}    \\
\rowcolor{gray!60} \textbf{RefineCoder-QW-7B} \ \ \textit{Iter3}   & 80K     &   \textbf{21.7}   &            \textbf{21.9}     & \textbf{20.8}      & 22.3      & \textbf{21.7}\textcolor[rgb]{0,0.5,0}{$_{\uparrow3.1}$}    \\
\bottomrule[1.5pt]
\end{tabular}
\caption{Results on LiveCodeBench and BigCodeBench-hard. v3-v5 and v5-v6 refer to the questions from the periods 202407-202502 and 202502-202505, respectively. \raisebox{-0.1cm}{\includegraphics[width=0.5cm]{figs/deepseek.png}} and \raisebox{-0.1cm}{\includegraphics[width=0.5cm]{figs/qwen.png}} denotes the two base models, and the rest are the instruction models. 2BT denotes 2B tokens.}
\label{exp:main}
\end{table*}

\section{Experiments}
\subsection{Benchmarks}

We use \textbf{HumanEval(+)}~\cite{chen2021evaluating}, \textbf{MBPP(+)}~\cite{austin2021programsynthesislargelanguage, liu2023is} and \textbf{BigCodeBench}~\cite{zhuo2024bigcodebench} to assess the \textbf{fundamental coding ability} of code model. Meanwhile, we use \textbf{LiveCodeBench}~\cite{jain2024livecodebench} and \textbf{BigCodeBench-hard} to measure the \textbf{advanced coding ability} of code model.

\subsection{Baselines}

\noindent $\circ$ \textbf{Based on DSCoder-6.7B-Base}: DSCoder-6.7B-Instruct~\cite{guo2024deepseek}, MagiCoder-S-DS-6.7B~\cite{wei2024magicoder}, OpenCodeInterpreter-DS-6.7B~\cite{zheng-etal-2024-opencodeinterpreter}, WaveCoder-Ultra-6.7B~\cite{yu-etal-2024-wavecoder}.

\noindent $\circ$ \textbf{Based on Qwen2.5-Coder-7B-Base}: Qwen2.5-Coder-7B-Instruct~\cite{hui2024qwen2}.

\subsection{Initial SFT Dataset $D_0$}

The existing datasets vary in quality and carry the risk of data leakage~\cite{wang-etal-2024-code, yu-etal-2024-wavecoder}. To ensure the accuracy and reliability of the experimental results, we construct a high-quality SFT dataset by combining Evol-Instruct and SelfCodeAlign~\cite{wei2024selfcodealign}. Concretely, we first prompt GPT-4o to generate code-related concepts and corresponding programming instructions from Python code snippets. These code snippets have been preprocessed to ensure high quality and contamination-free. Then, we prompt GPT-4o again to iteratively evolve these instructions using the Evol-Instruct strategy. Finally, we prompt GPT-4o to generate code responses for the instructions. Through this pipeline, we obtain an 80K high-quality and diverse python instruction dataset. See Appendix~\ref{appendix:sftdataset} for detailed data construction prompts.

\setlength{\tabcolsep}{4.5pt}
\begin{table*}[t]
\centering
\begin{tabular}{lccccccccll}
    \toprule[1.5pt]
    \multicolumn{1}{l}{\multirow{2}[2]{*}{\textbf{Models}}}  & \multicolumn{1}{l}{\multirow{2}[2]{*}{\textbf{Data Size}}} & \multicolumn{2}{c}{\textbf{HumanEval}} & \multicolumn{2}{c}{\textbf{MBPP}} & \multicolumn{2}{c}{\textbf{BigCodeBench}} & \multirow{2}[2]{*}{\textbf{Average}} \\
\cline{3-4} \cline{5-6} \cline{7-8}

& & \multicolumn{1}{c}{\textit{Base}} & \multicolumn{1}{c}{\textit{Plus}} & \multicolumn{1}{c}{\textit{Base}} & \multicolumn{1}{c}{\textit{Plus}} & \multicolumn{1}{c}{\textit{Inst}} & \multicolumn{1}{c}{\textit{Comp}} & \\    
    \midrule[1.5pt]
\raisebox{-0.1cm}{\includegraphics[width=0.5cm]{figs/deepseek.png}} DSCoder-6.7B-Base & / & 47.6 & 39.6 & 72.0 & 58.7 & / & 41.8 & /   \\
DSCoder-6.7B-Instruct & 2BT  &  74.4   & 71.3 & 74.9      & 65.6  & 35.5 & 43.8 & 60.9   \\
MagiCoder-S-DS-6.7B  & 185K  & \textcolor{red}{76.8} & \textcolor{red}{71.3} & \textcolor{red}{79.4}        & \textcolor{red}{\textbf{69.0}} & 36.2 & 47.6 & 63.4   \\
OpenCodeInterpreter-DS-6.7B & 178K &  \textcolor{red}{\textbf{77.4}}  & \textcolor{red}{\textbf{72.0}} & \textcolor{red}{76.5}    & \textcolor{red}{66.4}  & 37.1 & 44.6 & 62.3 \\
WaveCoder-Ultra-6.7B    & 130K & \textcolor{red}{75.0} & \textcolor{red}{69.5} & \textcolor{red}{74.9}      & \textcolor{red}{63.5}   &  33.9 & 43.7 & 60.1 \\
\rowcolor{gray!15} \textbf{RefineCoder-DS-6.7B} \ \ \textit{Iter0}       & 80K   & 73.8  &  67.7 & 77.5            & 65.1   &  37.0 & 46.6 & 61.3 \\
\rowcolor{gray!30}  \textbf{RefineCoder-DS-6.7B} \ \ \textit{Iter1}   & 80K     & 74.4 & 68.9           & 77.0    &    66.4     & 40.6 & 48.0 & 62.6\textcolor[rgb]{0,0.5,0}{$_{\uparrow1.3}$}                \\
\rowcolor{gray!45} \textbf{RefineCoder-DS-6.7B} \ \ \textit{Iter2}    & 80K          &  74.4  &  70.3         & 79.6             & 66.9          & 40.6 & 47.9 & 63.3\textcolor[rgb]{0,0.5,0}{$_{\uparrow2.0}$} \\
\rowcolor{gray!60} \textbf{RefineCoder-DS-6.7B} \ \ \textit{Iter3}   & 80K    & 75.0  & 70.7  & \textbf{80.2}      & 67.2 & \textbf{41.1} & \textbf{49.6} & \textbf{64.0}\textcolor[rgb]{0,0.5,0}{$_{\uparrow2.7}$}  \\
\midrule[1pt]
\raisebox{-0.1cm}{\includegraphics[width=0.5cm]{figs/qwen.png}}Qwen2.5-Coder-7B-Base  & /  & 61.6    & 53.0    & 76.9    & 62.9     & \ \ \ /  & \ \ \ / &  \ \ \ / \\
Qwen2.5-Coder-7B-Instruct & Millions & \textbf{88.4} & \textbf{84.1} & 83.5 & 71.7 & 40.4 & 48.8 & 69.5  \\
\rowcolor{gray!15} \textbf{RefineCoder-QW-7B} \ \ \textit{Iter0}       & 80K      &  84.9  &   78.1                    & 79.4            & 65.8   & 40.0 &  47.4  & 65.9 \\
\rowcolor{gray!30} \textbf{RefineCoder-QW-7B} \ \ \textit{Iter1}    & 80K   &   86.0 & 79.3                     & 84.1           & 70.8    & 41.6 & 49.0      & 68.5\textcolor[rgb]{0,0.5,0}{$_{ \uparrow2.6}$} \\
\rowcolor{gray!45} \textbf{RefineCoder-QW-7B} \ \ \textit{Iter2}   & 80K     &   86.6  &      82.5     & 83.4      & 71.7      & \textbf{42.3} & 50.9 & 69.6\textcolor[rgb]{0,0.5,0}{$_{\uparrow3.7}$}    \\
\rowcolor{gray!60} \textbf{RefineCoder-QW-7B} \ \ \textit{Iter3}   & 80K     &   87.2   &  83.3   & \textbf{85.6}      & \textbf{72.9}     & 42.0 & \textbf{51.1} & \textbf{70.4}\textcolor[rgb]{0,0.5,0}{$_{\uparrow4.5}$}    \\
\bottomrule[1.5pt]
\end{tabular}
\caption{Results on HumanEval, MBPP and BigCodeBench. Values in red indicate severe data leakage.}
\label{exp:main2}
\end{table*}

\setlength{\tabcolsep}{5pt}
\begin{table*}[t]
  \centering
  \small
    \begin{tabular}{lrrccccccc}
    \toprule[1.2pt]

    \textbf{Dataset} && \textbf{Size} && \textbf{HumanEval (+)} & \textbf{MBPP (+)} && \textbf{LCB} && \textbf{BCB} \\
    \midrule
    Magic-OSS-Instruct~\cite{wei2024magicoder} && 75K & & 4.45  &	9.40  &&	2.63 &&	4.56  \\
    Magic-Evol-Instruct~\cite{wei2024magicoder} && 110K & & \textcolor{red}{43.20} &	\textcolor{red}{19.40} &&	2.91 &&	4.10 \\
    Evol-CodeAlpaca-v1~\cite{luo2024wizardcoder}  && 110K & & \textcolor{red}{47.04} & \textcolor{red}{19.46}  &&	2.91 &&	4.69 \\
    Code-FeedBack~\cite{zheng-etal-2024-opencodeinterpreter} && 68K & & \textcolor{red}{30.50} &	\textcolor{red}{17.67} &&	3.16  && 4.49  \\
    \midrule
    Ours && 80K & & 4.97 	& 7.00 &&	1.54 &&	5.00  \\
    \bottomrule[1.2pt]
    \end{tabular}%
    \caption{The Test Leakage Indicator (TLI) quantifies data leakage by measuring the average maximum n-gram overlap between dataset samples and benchmark samples. The larger the TLI value, the more severe the data leakage.}
  \label{exp:dataleakage}%
\end{table*}%

\subsection{Implement Details}
\textbf{Iterative Training} We employ DSCoder-6.7B-Base and Qwen2.5-Coder-7B-Base as the base pre-trained models. To obtain an initial code model $M_0$, we fine-tune DSCoder-6.7B-Base and Qwen2.5-Coder-Base on the 80K initial SFT dataset, resulting in RefineCoder-DS-6.7B and RefineCoder-QW-7B. The two RefineCoder models undergo three iterative training under our ACR framework. In the ACR pipeline, the model self-samples 7 code responses with a temperature of 0.7. We use Qwen2.5-32B-Instruct~\cite{hui2024qwen2} as the Judge and Critic. We set the number of training epochs to 2, the global batch size to 64, and the learning rate to 5e-6, employing the AdamW optimizer with a cosine learning rate decay strategy. All our training uses 16 A100-80G GPUs, utilizing the LLaMA-Factory~\cite{zheng2024llamafactory}.

\textbf{Evaluation} We use the Pass@1 metric to evaluate the performance of model on benchmarks. We prioritize baseline results from the benchmark leaderboards or the original papers. If unavailable, we evaluate them locally using the same settings as the RefineCoder series.

\subsection{Experimental Results}
As shown in Table~\ref{exp:main}, the RefineCoder series achieves impressive performance gains on two challenging code benchmarks. While the initial RefineCoder-DS-6.7B and RefineCoder-QW-7B exhibit moderate performance, the two models achieve an average Pass@1 improvement of \textbf{2.7} and \textbf{3.1}, respectively, after three refinement iterations. Notably, the performance improved by \textbf{2.0} and \textbf{2.3} points from \textit{iter0} to \textit{iter1}, highlighting that incorporating self-generated code and external critique yields better results than directly distilling code responses from the teacher model. Compared to baselines of the same size, the RefineCoder-DS-6.7B and RefineCoder-QW-7B (\textit{Iter3}) outperform them by \textbf{3.6} and \textbf{2.3} points, respectively, using only 80K data. This demonstrates the superiority of our method in improving advanced coding ability.

Similarity, our method also improves the basic coding ability, as shown in Table~\ref{exp:main2}. After three iterations, our model achieves average performance improvements of \textbf{2.7} and \textbf{4.5} points, respectively, on HumanEval(+), MBPP(+) and BigCodeBench. This demonstrates that the model is capable of identifying and addressing deficiencies by learning from both its self-generated code and external feedback. Compared with all baselines, our models achieve the best average Pass@1 performance. Nevertheless, we find that the performance of RefineCoders on the Humaneval(+) are inferior to the state-of-the-art baseline. This discrepancy prompts us to conduct a data leakage analysis of the datasets in the next section.

\section{Further Study}

\subsection{Data Leakage Analysis}
In Table~\ref{exp:main2}, we observe that while the RefineCoder (\textit{Iter3}) surpasses the baselines on average pass@1, it still lags behind the state-of-the-art baseline on HumanEval(+) and MBPP+. To better explore this performance gap, we investigate potential data leakage in the datasets used by the baselines.

Concretely, we analyze the four open-source datasets~\footnote{\url{https://huggingface.co/datasets/ise-uiuc/Magicoder-OSS-Instruct-75K}}~\footnote{\url{https://huggingface.co/datasets/ise-uiuc/Magicoder-Evol-Instruct-110K}}~\footnote{\url{https://huggingface.co/datasets/theblackcat102/evol-codealpaca-v1}}~\footnote{\url{https://huggingface.co/datasets/m-a-p/Code-Feedback}} with the help of the TLI metric proposed by~\citet{wang-etal-2024-code}. These datasets are used by MagiCoder-S-DS-6.7B, OpenCodeInterpreter-DS-6.7B, and WaveCoder-Ultra-6.7B (The datasets for DSCoder-6.7B-Instruct and Qwen2.5-Coder-7B-Instruct are not publicly available). The results are shown in Table~\ref{exp:dataleakage}, Magic-OSS-Instruct and our dataset are entirely free from data leakage. In contrast, Magic-Evol-Instruct, Evol-CodeAlpaca-v1, and Code-FeedBack demonstrate severe data leakage concerning HumanEval (+) and MBPP (+). These three datasets are used by three baselines, making it challenging for RefineCoder to surpass them. We also present an intuitive scatter plot of similarity scores in the appendix~\ref{appendix:data_leakage} for a more detailed analysis.


\setlength{\tabcolsep}{2pt}
\begin{table}[t]
  \centering
  \small
    \begin{tabular}{lllcllc}
    \toprule
    \textbf{HumanEval+ (\%)} &&& \textbf{Best} && \textbf{Worst} & \textbf{Error Rate} \\
    \midrule
    RefineCoder-DS-6.7B \textit{Iter3} &&& 76.2  && 60.4\textcolor[rgb]{1,0,0}{$_{-15.8}$}  & 4.3   \\
    RefineCoder-QW-7B \textit{Iter3} &&& 83.6  && 75.0\textcolor[rgb]{1,0,0}{$_{-8.6}$}  & 3.1    \\
    \midrule
    \midrule
    \textbf{MBPP+ (\%)} &&& \textbf{Best} && \textbf{Worst} & \textbf{Error Rate} \\
    \midrule
    RefineCoder-DS-6.7B \textit{Iter3} &&& 70.3  && 60.1\textcolor[rgb]{1,0,0}{$_{-10.2}$}  & 5.8   \\
    RefineCoder-QW-7B \textit{Iter3} &&& 73.4  && 71.1\textcolor[rgb]{1,0,0}{$_{-2.3}$}  & 5.3    \\
    \bottomrule
    \end{tabular}%
    \caption{Best and Worst denote the performance of code with the highest and lowest scores. Error rate denotes the proportion of cases where the best code fails the test cases but the worst code passes.}
  \label{exp:score}%
\end{table}%

\subsection{The effectiveness of Composite Scoring System}
The performance improvement through iterative training preliminarily validates the effectiveness of our scoring system. In this section, we further verify it by enabling the model to sample 10 code responses for each programming question in benchmarks and scoring them. As shown in Table~\ref{exp:score}, the performance of the highest-scoring code far exceeds that of the lowest-scoring code, while the scoring system maintains a low error rate. This clearly validates its effectiveness.

\begin{figure}[t]
    \centering
    \includegraphics[width=1\linewidth]{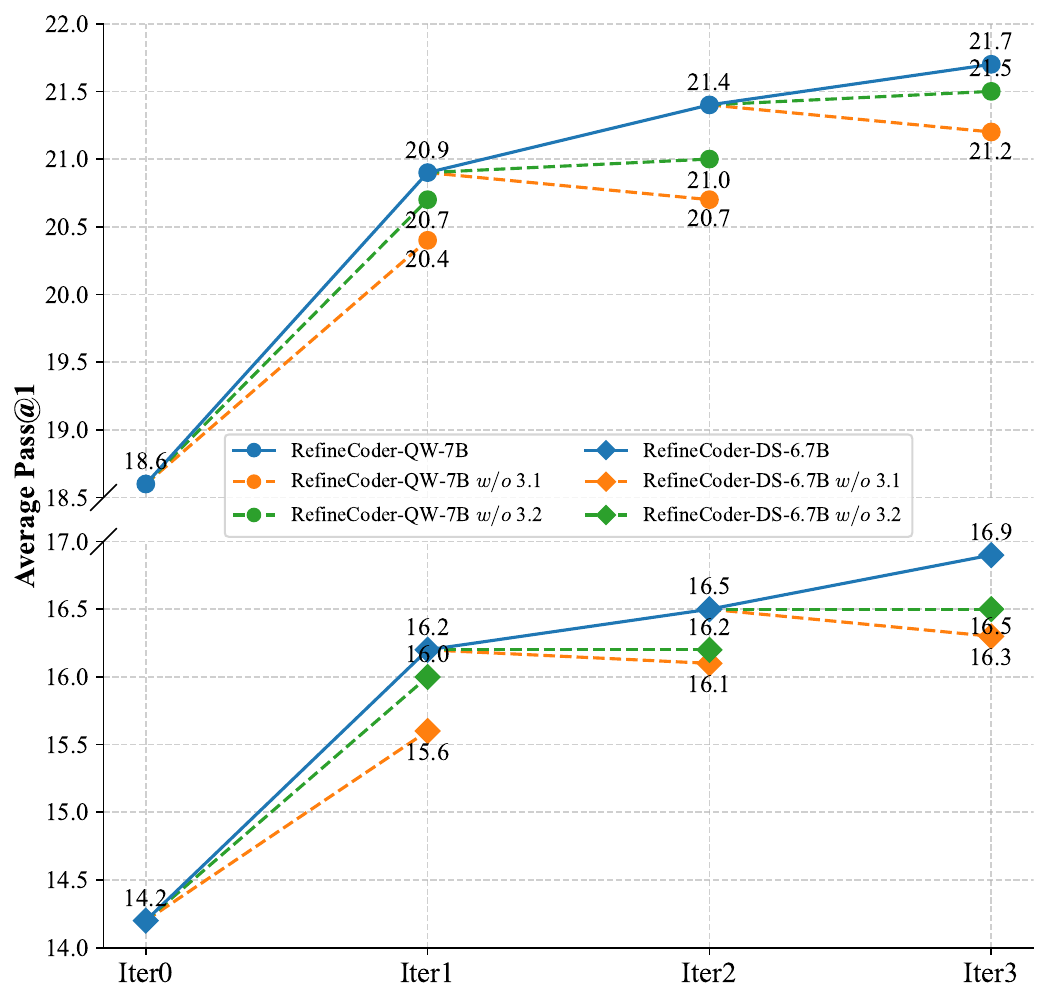}
    \caption{Ablation Study. The y-axis denotes the average pass@1 value on LiveCodeBench and BigCodeBench-hard.}
    \label{exp:ablation}
\end{figure}

\begin{figure*}[t]
    \centering
    \includegraphics[width=2.0\columnwidth]{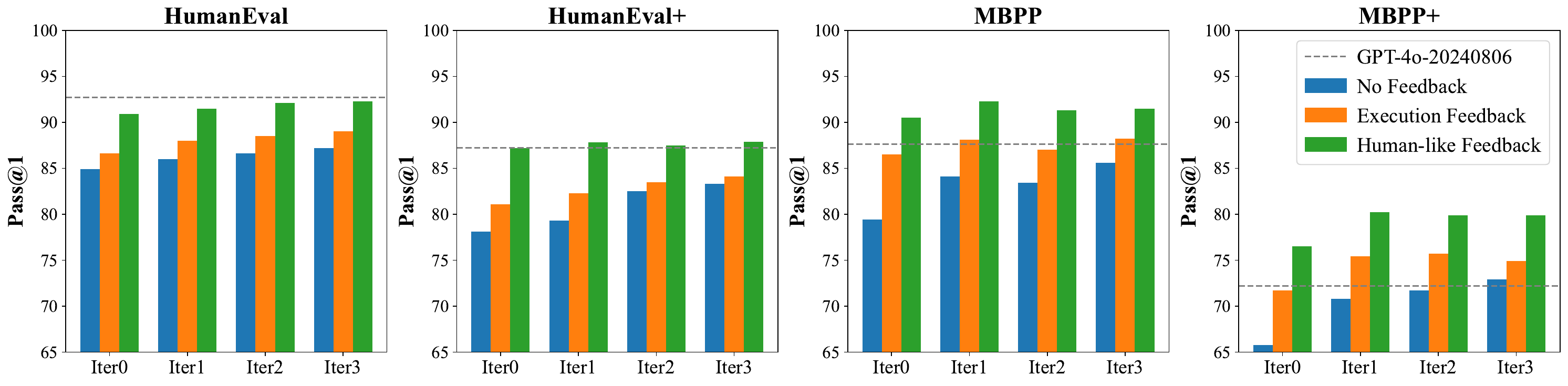}
    \caption{Evaluation with feedback using RefineCoder-QW-7B.}
    \label{exp:feedback}
\end{figure*}

\subsection{The effectiveness of Selective Critiquing Strategy}

We conduct ablation experiments to validate the effectiveness of the selective critiquing strategy in ACR. As shown in Figure~\ref{exp:ablation}, for each iteration, we construct two variants of RefineCoder (without 3.1 module of Figure~\ref{method:model} or without 3.2 module). During the iteration process, ablating either module leads to a performance drop in both RefineCoder-QW-7B and RefineCoder-DS-6.7B, indicating the effectiveness of our strategy. Furthermore, removing the second-turn critique data has a greater impact on performance. We believe this action prevents the model from reflecting on self-generated low-quality code and external critique, thereby hindering performance improvement.

\subsection{Multilingual Code Generation}

\setlength{\tabcolsep}{4.5pt}
\begin{table*}[t]
\small
  \centering
    \begin{tabular}{lccccccccccc}
    \toprule
          \textbf{Models} & & \textbf{C++} && \textbf{C\#} & &\textbf{Java} && \textbf{Bash} & \textbf{TypeScript} & \textbf{JavaScript} & \textbf{AVG} \\
    \midrule
    \textbf{RefineCoder-DS-6.7B} \ \ \textit{Iter0} & & 56.5  && 58.9  && \textbf{58.9} && \textbf{40.5} & 62.3  & 64.6  & 57.2  \\
    \textbf{RefineCoder-DS-6.7B} \ \ \textit{Iter1}  & & 62.1  && \textbf{61.4} && 55.1  && 35.4  & \textbf{65.4} & \textbf{66.5} & \textbf{58.2}  \\
    \textbf{RefineCoder-DS-6.7B} \ \ \textit{Iter2}  & & \textbf{63.4} && 56.2  && 56.2  && 38.6  & \textbf{65.4} & 65.8  & 57.5  \\
    \textbf{RefineCoder-DS-6.7B} \ \ \textit{Iter3}  & & \textbf{63.4} && 58.2  && 57.0  && 39.2  & 62.9  & 65.8  & 57.8  \\
    \midrule
    \textbf{RefineCoder-QW-7B} \ \ \textit{Iter0} & & 60.7  && 72.4 & & 68.0 & & 49.9  & 77.6  & \textbf{78.1} &   67.8 \\
    \textbf{RefineCoder-QW-7B} \ \ \textit{Iter1} & & 62.1  && 74.6 & & 65.0  && 50.8  & 76.8  & 77.4  &   67.8 \\
    \textbf{RefineCoder-QW-7B} \ \ \textit{Iter2} & & \textbf{63.7} && 74.1 & & 69.6 & & 49.7  & 78.3    & 77.7  &  68.9 \\
    \textbf{RefineCoder-QW-7B} \ \ \textit{Iter3} & & 60.9  && \textbf{75.3} && \textbf{73.8} && \textbf{52.2} & \textbf{79.3} & 77.9  &  \textbf{69.9} \\
    \bottomrule
    \end{tabular}%
     \caption{Performance of RefineCoder on the MultiPL-E.}
  \label{exp:multiple}%
\end{table*}%

As shown in Table~\ref{exp:multiple}, we also evaluate the out-of-distribution (OOD) code generation performance on the multilingual benchmark MultiPL-E~\cite{cassano2022multipl} despite fine-tuning on a Python-only dataset. After one and three iterations, RefineCoder-DS-6.7B and RefineCoder-QW-7B achieve optimal average performance, surpassing the initial model by 1.0 and 2.1 points, respectively. In particular, RefineCoder-QW-7B (\textit{Iter3}) achieves the best results among half of the programming languages. This demonstrates that the effectiveness of our ACR method generalizes well to the OOD code generation domain.

\subsection{Evaluation with External Feedback}

Iterative ACR not only improves the one-off code generation performance of RefineCoder but also endows it with the ability to correct errors based on feedback. Following ~\citet{zheng-etal-2024-opencodeinterpreter}, we design two types of external feedback to evaluate this ability of RefineCoder: 1) \textbf{Execution Feedback}: Model refines its self-generated erroneous code based on execution results from the executor; 2) \textbf{Human-like Feedback}: GPT-4o first analyzes the programming question, initial error code, and execution results to generate improvement suggestions that mimic human thinking. The model then refines the code based on these suggestions. The results of RefineCoder-QW-7B on HumanEval(+) and MBPP(+) are shown in Figure~\ref{exp:feedback}. It can be observed that external feedback improves performance on all benchmarks, with human-like feedback yielding a more significant enhancement. Surprisingly, with the help of human-like improvement suggestions, RefineCoder-QW-7B (\textit{Iter1} and \textit{Iter3}) outperforms GPT-4o on three benchmarks, demonstrating its ability to understand feedback and refine code.

\section{Conclusions}

In this paper, we propose the Adaptive Critique Refinement (ACR) method to iteratively refine code LLMs using self-generated code and external critique, which breaks away from the traditional teacher distillation paradigm and improves the intrinsic one-off code generation ability. We design a composite scoring system and a selective critiquing strategy. These two components are centered around LLM-as-a-Judge and LLM-as-a-Critic to evaluate and critique self-generated code. This simulates the process where a student solves a problem independently and then refines it by comparing it with the reference answer. We develop the RefineCoder-DS-6.7B and RefineCoder-QW-7B models and demonstrate the effectiveness of iterative ACR on multiple code generation benchmarks. Further studies reveal the impact of the each component in ACR.

\section{Limitations}

Although Adaptive Critique Refinement (ACR) and the associated RefineCoder demonstrate significant improvements in the code generation task, several limitations remain. First, ACR is primarily designed to refine code responses based on programming instructions, necessitating an initial set of high-quality and diverse instructions. Therefore, a specialized code instruction generator is still required to make ACR more automated. Furthermore, while ACR can apply to other reasoning-intensive tasks, such as mathematics, this paper has not fully explored these domains.


\renewcommand{\ULthickness}{0pt}
\renewcommand{\uline}{\relax}

\bibliography{custom}

\begin{thebibliography}{47}
\providecommand{\natexlab}[1]{#1}

\bibitem[{Anthropic(2024)}]{claude3p5-blog}
Anthropic. 2024.
\newblock Introducing computer use, a new claude 3.5 sonnet, and claude 3.5 haiku.
\newblock \url{https://www.anthropic.com/news/3-5-models-and-computer-use}.

\bibitem[{Austin et~al.(2021)Austin, Odena, Nye, Bosma, Michalewski, Dohan, Jiang, Cai, Terry, Le, and Sutton}]{austin2021programsynthesislargelanguage}
Jacob Austin, Augustus Odena, Maxwell Nye, Maarten Bosma, Henryk Michalewski, David Dohan, Ellen Jiang, Carrie Cai, Michael Terry, Quoc Le, and Charles Sutton. 2021.
\newblock \href {https://arxiv.org/abs/2108.07732} {Program synthesis with large language models}.
\newblock \emph{Preprint}, arXiv:2108.07732.

\bibitem[{Cassano et~al.(2022)Cassano, Gouwar, Nguyen, Nguyen, Phipps-Costin, Pinckney, Yee, Zi, Anderson, Feldman et~al.}]{cassano2022multipl}
Federico Cassano, John Gouwar, Daniel Nguyen, Sydney Nguyen, Luna Phipps-Costin, Donald Pinckney, Ming-Ho Yee, Yangtian Zi, Carolyn~Jane Anderson, Molly~Q Feldman, et~al. 2022.
\newblock Multipl-e: A scalable and extensible approach to benchmarking neural code generation.
\newblock \emph{arXiv preprint arXiv:2208.08227}.

\bibitem[{Chaudhary(2023)}]{codealpaca}
Sahil Chaudhary. 2023.
\newblock Code alpaca: An instruction-following llama model for code generation.
\newblock \url{https://github.com/sahil280114/codealpaca}.

\bibitem[{Chen et~al.(2023)Chen, Zhang, Nguyen, Zan, Lin, Lou, and Chen}]{chen2023codet}
Bei Chen, Fengji Zhang, Anh Nguyen, Daoguang Zan, Zeqi Lin, Jian-Guang Lou, and Weizhu Chen. 2023.
\newblock \href {https://openreview.net/forum?id=ktrw68Cmu9c} {Codet: Code generation with generated tests}.
\newblock In \emph{The Eleventh International Conference on Learning Representations}.

\bibitem[{Chen et~al.(2021)Chen, Tworek, Jun, Yuan, Pinto, Kaplan, Edwards, Burda, Joseph, Brockman et~al.}]{chen2021evaluating}
Mark Chen, Jerry Tworek, Heewoo Jun, Qiming Yuan, Henrique Ponde De~Oliveira Pinto, Jared Kaplan, Harri Edwards, Yuri Burda, Nicholas Joseph, Greg Brockman, et~al. 2021.
\newblock Evaluating large language models trained on code.
\newblock \emph{arXiv preprint arXiv:2107.03374}.

\bibitem[{Chen et~al.(2025)Chen, Tao, Zhang, Zhou, Gu, He, Zhang, Cai, Zhao, and Jin}]{chen2025revisitselfdebuggingselfgeneratedtests}
Xiancai Chen, Zhengwei Tao, Kechi Zhang, Changzhi Zhou, Wanli Gu, Yuanpeng He, Mengdi Zhang, Xunliang Cai, Haiyan Zhao, and Zhi Jin. 2025.
\newblock \href {https://arxiv.org/abs/2501.12793} {Revisit self-debugging with self-generated tests for code generation}.
\newblock \emph{Preprint}, arXiv:2501.12793.

\bibitem[{Chen et~al.(2024)Chen, Lin, Sch{\"a}rli, and Zhou}]{chen2024teaching}
Xinyun Chen, Maxwell Lin, Nathanael Sch{\"a}rli, and Denny Zhou. 2024.
\newblock \href {https://openreview.net/forum?id=KuPixIqPiq} {Teaching large language models to self-debug}.
\newblock In \emph{The Twelfth International Conference on Learning Representations}.

\bibitem[{DeepSeek-AI(2025)}]{2025deepseekr1}
DeepSeek-AI. 2025.
\newblock \href {https://arxiv.org/abs/2501.12948} {Deepseek-r1: Incentivizing reasoning capability in llms via reinforcement learning}.
\newblock \emph{Preprint}, arXiv:2501.12948.

\bibitem[{Dong et~al.(2025)Dong, Dong, Zhang, Sui, and Wei}]{dong2024self}
Qingxiu Dong, Li~Dong, Xingxing Zhang, Zhifang Sui, and Furu Wei. 2025.
\newblock \href {https://openreview.net/forum?id=7visV100Ms} {Self-boosting large language models with synthetic preference data}.
\newblock In \emph{The Thirteenth International Conference on Learning Representations}.

\bibitem[{Google(2024)}]{gemini-blog}
Google. 2024.
\newblock Gemini.
\newblock \url{https://deepmind.google/technologies/gemini/}.

\bibitem[{Gulwani et~al.(2017)Gulwani, Polozov, Singh et~al.}]{gulwani2017program}
Sumit Gulwani, Oleksandr Polozov, Rishabh Singh, et~al. 2017.
\newblock Program synthesis.
\newblock \emph{Foundations and Trends{\textregistered} in Programming Languages}, 4(1-2):1--119.

\bibitem[{Guo et~al.(2024)Guo, Zhu, Yang, Xie, Dong, Zhang, Chen, Bi, Wu, Li et~al.}]{guo2024deepseek}
Daya Guo, Qihao Zhu, Dejian Yang, Zhenda Xie, Kai Dong, Wentao Zhang, Guanting Chen, Xiao Bi, Yu~Wu, YK~Li, et~al. 2024.
\newblock Deepseek-coder: When the large language model meets programming--the rise of code intelligence.
\newblock \emph{arXiv preprint arXiv:2401.14196}.

\bibitem[{Hu et~al.(2024)Hu, Hu, Cao, Xiao, and Zhu}]{hu-etal-2024-teaching}
Chi Hu, Yimin Hu, Hang Cao, Tong Xiao, and JingBo Zhu. 2024.
\newblock \href {https://doi.org/10.18653/v1/2024.findings-acl.364} {Teaching language models to self-improve by learning from language feedback}.
\newblock In \emph{Findings of the Association for Computational Linguistics: ACL 2024}, pages 6090--6101, Bangkok, Thailand. Association for Computational Linguistics.

\bibitem[{Huang et~al.(2023)Huang, Gu, Hou, Wu, Wang, Yu, and Han}]{huang-etal-2023-large}
Jiaxin Huang, Shixiang Gu, Le~Hou, Yuexin Wu, Xuezhi Wang, Hongkun Yu, and Jiawei Han. 2023.
\newblock \href {https://doi.org/10.18653/v1/2023.emnlp-main.67} {Large language models can self-improve}.
\newblock In \emph{Proceedings of the 2023 Conference on Empirical Methods in Natural Language Processing}, pages 1051--1068, Singapore. Association for Computational Linguistics.

\bibitem[{Hui et~al.(2024)Hui, Yang, Cui, Yang, Liu, Zhang, Liu, Zhang, Yu, Dang et~al.}]{hui2024qwen2}
Binyuan Hui, Jian Yang, Zeyu Cui, Jiaxi Yang, Dayiheng Liu, Lei Zhang, Tianyu Liu, Jiajun Zhang, Bowen Yu, Kai Dang, et~al. 2024.
\newblock Qwen2.5-coder technical report.
\newblock \emph{arXiv preprint arXiv:2409.12186}.

\bibitem[{Jain et~al.(2024)Jain, Han, Gu, Li, Yan, Zhang, Wang, Solar-Lezama, Sen, and Stoica}]{jain2024livecodebench}
Naman Jain, King Han, Alex Gu, Wen-Ding Li, Fanjia Yan, Tianjun Zhang, Sida Wang, Armando Solar-Lezama, Koushik Sen, and Ion Stoica. 2024.
\newblock Livecodebench: Holistic and contamination free evaluation of large language models for code.
\newblock \emph{arXiv preprint arXiv:2403.07974}.

\bibitem[{Jiang et~al.(2025)Jiang, Yan, Liu, Jin, Peng, Zhang, Cai, Cao, Gao, and Tang}]{jiang2025logicproimprovingcomplexlogical}
Jin Jiang, Yuchen Yan, Yang Liu, Yonggang Jin, Shuai Peng, Mengdi Zhang, Xunliang Cai, Yixin Cao, Liangcai Gao, and Zhi Tang. 2025.
\newblock \href {https://arxiv.org/abs/2409.12929} {Logicpro: Improving complex logical reasoning via program-guided learning}.
\newblock \emph{Preprint}, arXiv:2409.12929.

\bibitem[{Kim et~al.(2025)Kim, Kim, Lee, and Shin}]{kim2025spread}
Dongyoung Kim, Jaehyung Kim, Kimin Lee, and Jinwoo Shin. 2025.
\newblock \href {https://openreview.net/forum?id=BPgK5XW1Nb} {Spread preference annotation: Direct preference judgment for efficient {LLM} alignment}.
\newblock In \emph{The Thirteenth International Conference on Learning Representations}.

\bibitem[{Kocetkov et~al.(2022)Kocetkov, Li, Allal, Li, Mou, Ferrandis, Jernite, Mitchell, Hughes, Wolf et~al.}]{kocetkov2022stack}
Denis Kocetkov, Raymond Li, Loubna~Ben Allal, Jia Li, Chenghao Mou, Carlos~Mu{\~n}oz Ferrandis, Yacine Jernite, Margaret Mitchell, Sean Hughes, Thomas Wolf, et~al. 2022.
\newblock The stack: 3 tb of permissively licensed source code.
\newblock \emph{arXiv preprint arXiv:2211.15533}.

\bibitem[{Li et~al.(2024)Li, Jiang, Huang, Beigi, Zhao, Tan, Bhattacharjee, Jiang, Chen, Wu, Shu, Cheng, and Liu}]{li2024llmasajudge}
Dawei Li, Bohan Jiang, Liangjie Huang, Alimohammad Beigi, Chengshuai Zhao, Zhen Tan, Amrita Bhattacharjee, Yuxuan Jiang, Canyu Chen, Tianhao Wu, Kai Shu, Lu~Cheng, and Huan Liu. 2024.
\newblock From generation to judgment: Opportunities and challenges of llm-as-a-judge.
\newblock \emph{arXiv preprint arXiv: 2411.16594}.

\bibitem[{Liu et~al.(2023)Liu, Xia, Wang, and ZHANG}]{liu2023is}
Jiawei Liu, Chunqiu~Steven Xia, Yuyao Wang, and LINGMING ZHANG. 2023.
\newblock \href {https://openreview.net/forum?id=1qvx610Cu7} {Is your code generated by chat{GPT} really correct? rigorous evaluation of large language models for code generation}.
\newblock In \emph{Thirty-seventh Conference on Neural Information Processing Systems}.

\bibitem[{Luo et~al.(2024)Luo, Xu, Zhao, Sun, Geng, Hu, Tao, Ma, Lin, and Jiang}]{luo2024wizardcoder}
Ziyang Luo, Can Xu, Pu~Zhao, Qingfeng Sun, Xiubo Geng, Wenxiang Hu, Chongyang Tao, Jing Ma, Qingwei Lin, and Daxin Jiang. 2024.
\newblock \href {https://openreview.net/forum?id=UnUwSIgK5W} {Wizardcoder: Empowering code large language models with evol-instruct}.
\newblock In \emph{The Twelfth International Conference on Learning Representations}.

\bibitem[{Madaan et~al.(2023)Madaan, Tandon, Gupta, Hallinan, Gao, Wiegreffe, Alon, Dziri, Prabhumoye, Yang, Gupta, Majumder, Hermann, Welleck, Yazdanbakhsh, and Clark}]{madaan2023selfrefine}
Aman Madaan, Niket Tandon, Prakhar Gupta, Skyler Hallinan, Luyu Gao, Sarah Wiegreffe, Uri Alon, Nouha Dziri, Shrimai Prabhumoye, Yiming Yang, Shashank Gupta, Bodhisattwa~Prasad Majumder, Katherine Hermann, Sean Welleck, Amir Yazdanbakhsh, and Peter Clark. 2023.
\newblock \href {https://openreview.net/forum?id=S37hOerQLB} {Self-refine: Iterative refinement with self-feedback}.
\newblock In \emph{Thirty-seventh Conference on Neural Information Processing Systems}.

\bibitem[{OpenAI(2024{\natexlab{a}})}]{gpt4o-blog}
OpenAI. 2024{\natexlab{a}}.
\newblock Hello gpt-4o.
\newblock \url{https://openai.com/index/hello-gpt-4o/}.

\bibitem[{OpenAI(2024{\natexlab{b}})}]{o1-blog}
OpenAI. 2024{\natexlab{b}}.
\newblock Learning to reason with llms.
\newblock \url{https://openai.com/index/learning-to-reason-with-llms/}.

\bibitem[{Quan et~al.(2025)Quan, Yang, Yu, Zheng, Liu, Yang, Ren, Gao, Miao, Feng et~al.}]{quan2025codeelo}
Shanghaoran Quan, Jiaxi Yang, Bowen Yu, Bo~Zheng, Dayiheng Liu, An~Yang, Xuancheng Ren, Bofei Gao, Yibo Miao, Yunlong Feng, et~al. 2025.
\newblock Codeelo: Benchmarking competition-level code generation of llms with human-comparable elo ratings.
\newblock \emph{arXiv preprint arXiv:2501.01257}.

\bibitem[{Rafailov et~al.(2023)Rafailov, Sharma, Mitchell, Manning, Ermon, and Finn}]{rafailov2023direct}
Rafael Rafailov, Archit Sharma, Eric Mitchell, Christopher~D Manning, Stefano Ermon, and Chelsea Finn. 2023.
\newblock \href {https://openreview.net/forum?id=HPuSIXJaa9} {Direct preference optimization: Your language model is secretly a reward model}.
\newblock In \emph{Thirty-seventh Conference on Neural Information Processing Systems}.

\bibitem[{Roziere et~al.(2023)Roziere, Gehring, Gloeckle, Sootla, Gat, Tan, Adi, Liu, Sauvestre, Remez et~al.}]{roziere2023code}
Baptiste Roziere, Jonas Gehring, Fabian Gloeckle, Sten Sootla, Itai Gat, Xiaoqing~Ellen Tan, Yossi Adi, Jingyu Liu, Romain Sauvestre, Tal Remez, et~al. 2023.
\newblock Code llama: Open foundation models for code.
\newblock \emph{arXiv preprint arXiv:2308.12950}.

\bibitem[{Tao et~al.(2024)Tao, Chen, Yu, Mai, Rossi, Li, and Mitra}]{tao2024codelutra}
Leitian Tao, Xiang Chen, Tong Yu, Tung Mai, Ryan Rossi, Yixuan Li, and Saayan Mitra. 2024.
\newblock Codelutra: Boosting llm code generation via preference-guided refinement.
\newblock \emph{arXiv preprint arXiv:2411.05199}.

\bibitem[{Wang et~al.(2024{\natexlab{a}})Wang, Kulikov, Golovneva, Yu, Yuan, Dwivedi-Yu, Pang, Fazel-Zarandi, Weston, and Li}]{wang2024selftaughtevaluators}
Tianlu Wang, Ilia Kulikov, Olga Golovneva, Ping Yu, Weizhe Yuan, Jane Dwivedi-Yu, Richard~Yuanzhe Pang, Maryam Fazel-Zarandi, Jason Weston, and Xian Li. 2024{\natexlab{a}}.
\newblock \href {https://arxiv.org/abs/2408.02666} {Self-taught evaluators}.
\newblock \emph{Preprint}, arXiv:2408.02666.

\bibitem[{Wang et~al.(2024{\natexlab{b}})Wang, He, Fu, GongQue, Xu, Chen, Wang, Fu, Dong, Diao, Wang, Zhang, Cai, and Xu}]{wang-etal-2024-code}
Yejie Wang, Keqing He, Dayuan Fu, Zhuoma GongQue, Heyang Xu, Yanxu Chen, Zhexu Wang, Yujia Fu, Guanting Dong, Muxi Diao, Jingang Wang, Mengdi Zhang, Xunliang Cai, and Weiran Xu. 2024{\natexlab{b}}.
\newblock \href {https://doi.org/10.18653/v1/2024.emnlp-main.777} {How do your code {LLM}s perform? empowering code instruction tuning with really good data}.
\newblock In \emph{Proceedings of the 2024 Conference on Empirical Methods in Natural Language Processing}, pages 14027--14043, Miami, Florida, USA. Association for Computational Linguistics.

\bibitem[{Wang et~al.(2023)Wang, Kordi, Mishra, Liu, Smith, Khashabi, and Hajishirzi}]{wang-etal-2023-self-instruct}
Yizhong Wang, Yeganeh Kordi, Swaroop Mishra, Alisa Liu, Noah~A. Smith, Daniel Khashabi, and Hannaneh Hajishirzi. 2023.
\newblock \href {https://doi.org/10.18653/v1/2023.acl-long.754} {Self-instruct: Aligning language models with self-generated instructions}.
\newblock In \emph{Proceedings of the 61st Annual Meeting of the Association for Computational Linguistics (Volume 1: Long Papers)}, pages 13484--13508, Toronto, Canada. Association for Computational Linguistics.

\bibitem[{Wang et~al.(2025)Wang, Yue, and Chen}]{wang2025critique}
Yubo Wang, Xiang Yue, and Wenhu Chen. 2025.
\newblock Critique fine-tuning: Learning to critique is more effective than learning to imitate.
\newblock \emph{arXiv preprint arXiv:2501.17703}.

\bibitem[{Wei et~al.(2024{\natexlab{a}})Wei, Cassano, Liu, Ding, Jain, Mueller, de~Vries, Werra, Guha, and ZHANG}]{wei2024selfcodealign}
Yuxiang Wei, Federico Cassano, Jiawei Liu, Yifeng Ding, Naman Jain, Zachary Mueller, Harm de~Vries, Leandro~Von Werra, Arjun Guha, and LINGMING ZHANG. 2024{\natexlab{a}}.
\newblock \href {https://openreview.net/forum?id=xXRnUU7xTL} {Selfcodealign: Self-alignment for code generation}.
\newblock In \emph{The Thirty-eighth Annual Conference on Neural Information Processing Systems}.

\bibitem[{Wei et~al.(2024{\natexlab{b}})Wei, Wang, Liu, Ding, and ZHANG}]{wei2024magicoder}
Yuxiang Wei, Zhe Wang, Jiawei Liu, Yifeng Ding, and LINGMING ZHANG. 2024{\natexlab{b}}.
\newblock \href {https://openreview.net/forum?id=XUeoOBid3x} {Magicoder: Empowering code generation with {OSS}-instruct}.
\newblock In \emph{Forty-first International Conference on Machine Learning}.

\bibitem[{Wu et~al.(2024)Wu, Huang, Shi, Wang, Gao, Liu, Nan, Yuan, Zhang, Zhang, Du, Guo, Pu, Yin, Hu, and Chen}]{wu2024inversecoders}
Yutong Wu, Di~Huang, Wenxuan Shi, Wei Wang, Lingzhe Gao, Shihao Liu, Ziyuan Nan, Kaizhao Yuan, Rui Zhang, Xishan Zhang, Zidong Du, Qi~Guo, Yewen Pu, Dawei Yin, Xing Hu, and Yunji Chen. 2024.
\newblock \href {https://arxiv.org/abs/2407.05700} {Inversecoder: Self-improving instruction-tuned code llms with inverse-instruct}.
\newblock \emph{Preprint}, arXiv:2407.05700.

\bibitem[{Xu et~al.(2024)Xu, Sun, Zheng, Geng, Zhao, Feng, Tao, Lin, and Jiang}]{xu2024wizardlm}
Can Xu, Qingfeng Sun, Kai Zheng, Xiubo Geng, Pu~Zhao, Jiazhan Feng, Chongyang Tao, Qingwei Lin, and Daxin Jiang. 2024.
\newblock \href {https://openreview.net/forum?id=CfXh93NDgH} {Wizard{LM}: Empowering large pre-trained language models to follow complex instructions}.
\newblock In \emph{The Twelfth International Conference on Learning Representations}.

\bibitem[{Yan et~al.(2025)Yan, Jiang, Liu, Cao, Xu, Zhang, Cai, and Shao}]{s3cmath}
Yuchen Yan, Jin Jiang, Yang Liu, Yixin Cao, Xin Xu, Mengdi Zhang, Xunliang Cai, and Jian Shao. 2025.
\newblock \href {https://doi.org/10.1609/aaai.v39i24.34749} {S3cmath: Spontaneous step-level self-correction makes large language models better mathematical reasoners}.
\newblock \emph{Proceedings of the AAAI Conference on Artificial Intelligence}, 39(24):25588--25596.

\bibitem[{Yu et~al.(2024)Yu, Zhang, Shang, Huang, Xu, Zhao, Hu, and Yin}]{yu-etal-2024-wavecoder}
Zhaojian Yu, Xin Zhang, Ning Shang, Yangyu Huang, Can Xu, Yishujie Zhao, Wenxiang Hu, and Qiufeng Yin. 2024.
\newblock \href {https://doi.org/10.18653/v1/2024.acl-long.280} {{W}ave{C}oder: Widespread and versatile enhancement for code large language models by instruction tuning}.
\newblock In \emph{Proceedings of the 62nd Annual Meeting of the Association for Computational Linguistics (Volume 1: Long Papers)}, pages 5140--5153, Bangkok, Thailand. Association for Computational Linguistics.

\bibitem[{Yuan et~al.(2024)Yuan, Pang, Cho, Li, Sukhbaatar, Xu, and Weston}]{yuan2024selfrewarding}
Weizhe Yuan, Richard~Yuanzhe Pang, Kyunghyun Cho, Xian Li, Sainbayar Sukhbaatar, Jing Xu, and Jason~E Weston. 2024.
\newblock \href {https://openreview.net/forum?id=0NphYCmgua} {Self-rewarding language models}.
\newblock In \emph{Forty-first International Conference on Machine Learning}.

\bibitem[{Zhang et~al.(2024{\natexlab{a}})Zhang, Li, Dong, Xu, Zhang, Su, Liu, and Jin}]{zhang2024codedpo}
Kechi Zhang, Ge~Li, Yihong Dong, Jingjing Xu, Jun Zhang, Jing Su, Yongfei Liu, and Zhi Jin. 2024{\natexlab{a}}.
\newblock Codedpo: Aligning code models with self generated and verified source code.
\newblock \emph{arXiv preprint arXiv:2410.05605}.

\bibitem[{Zhang et~al.(2024{\natexlab{b}})Zhang, Chen, Liu, Liao, Gong, Yu, Li, and Wang}]{zhang2024unifyingpe}
Ziyin Zhang, Chaoyu Chen, Bingchang Liu, Cong Liao, Zi~Gong, Hang Yu, Jianguo Li, and Rui Wang. 2024{\natexlab{b}}.
\newblock \href {https://arxiv.org/abs/2311.07989} {Unifying the perspectives of nlp and software engineering: A survey on language models for code}.
\newblock \emph{Preprint}, arXiv:2311.07989.

\bibitem[{Zheng et~al.(2024{\natexlab{a}})Zheng, Zhang, Shen, Liu, Lin, Fu, Chen, and Yue}]{zheng-etal-2024-opencodeinterpreter}
Tianyu Zheng, Ge~Zhang, Tianhao Shen, Xueling Liu, Bill~Yuchen Lin, Jie Fu, Wenhu Chen, and Xiang Yue. 2024{\natexlab{a}}.
\newblock \href {https://doi.org/10.18653/v1/2024.findings-acl.762} {{O}pen{C}ode{I}nterpreter: Integrating code generation with execution and refinement}.
\newblock In \emph{Findings of the Association for Computational Linguistics: ACL 2024}, pages 12834--12859, Bangkok, Thailand. Association for Computational Linguistics.

\bibitem[{Zheng et~al.(2024{\natexlab{b}})Zheng, Zhang, Zhang, Ye, Luo, Feng, and Ma}]{zheng2024llamafactory}
Yaowei Zheng, Richong Zhang, Junhao Zhang, Yanhan Ye, Zheyan Luo, Zhangchi Feng, and Yongqiang Ma. 2024{\natexlab{b}}.
\newblock \href {http://arxiv.org/abs/2403.13372} {Llamafactory: Unified efficient fine-tuning of 100+ language models}.
\newblock In \emph{Proceedings of the 62nd Annual Meeting of the Association for Computational Linguistics (Volume 3: System Demonstrations)}, Bangkok, Thailand. Association for Computational Linguistics.

\bibitem[{Zhong et~al.(2024)Zhong, Wang, and Shang}]{zhong-etal-2024-debug}
Li~Zhong, Zilong Wang, and Jingbo Shang. 2024.
\newblock \href {https://doi.org/10.18653/v1/2024.findings-acl.49} {Debug like a human: A large language model debugger via verifying runtime execution step by step}.
\newblock In \emph{Findings of the Association for Computational Linguistics: ACL 2024}, pages 851--870, Bangkok, Thailand. Association for Computational Linguistics.

\bibitem[{Zhuo et~al.(2024)Zhuo, Vu, Chim, Hu, Yu, Widyasari, Yusuf, Zhan, He, Paul et~al.}]{zhuo2024bigcodebench}
Terry~Yue Zhuo, Minh~Chien Vu, Jenny Chim, Han Hu, Wenhao Yu, Ratnadira Widyasari, Imam Nur~Bani Yusuf, Haolan Zhan, Junda He, Indraneil Paul, et~al. 2024.
\newblock Bigcodebench: Benchmarking code generation with diverse function calls and complex instructions.
\newblock \emph{arXiv preprint arXiv:2406.15877}.

\end{thebibliography}

\renewcommand{\ULthickness}{0.8pt}
\renewcommand{\uline}{\ULine}

\appendix

\newpage








\section{Prompts for Judge and Critic}
\label{appendix:judge_critic}

The prompt for point/pair-wise LLM-as-a-Judge and LLM-as-a-Critic are shown in Figure~\ref{prompt:pointjudge}, ~\ref{prompt:pairjudge} and ~\ref{prompt:paircritic}.

\section{Prompts for Constructing SFT Dataset}
\label{appendix:sftdataset}

We called GPT-4o to create the SFT dataset, with the following prompts used:

$\circ$ Prompt for generating code-related concepts and programming instructions, as shown in Figure~\ref{prompt:gen_ins}.

$\circ$ Prompts for evolving existing instructions, as shown in Figures~\ref{prompt:evol_addition}, ~\ref{prompt:evol_breath}, ~\ref{prompt:evol_comp},~\ref{prompt:evol_concre},~\ref{prompt:evol_deep},~\ref{prompt:evol_diversion},~\ref{prompt:evol_increase}, ~\ref{prompt:evol_misd}, ~\ref{prompt:evol_reason}.

$\circ$ For generating code responses, We directly feed the instructions to the model.

\section{Prompts for Evaluation}
\label{appendix:eval}

For Humaneval (+) and MBPP (+), we use the prompts designed by OpenCodeInterpreter~\cite{zheng-etal-2024-opencodeinterpreter}; for Livecodebench and Bigcodebench-hard, we use the official prompts. When conducting evaluations with execution feedback, we have designed a prompt as illustrated in Figure~\ref{prompt:eval_exec}. When executing human-like feedback, following OpenCodeInterpreter, we generate two prompts: one for generating improvement suggestions and the other for refine code according to human-like feedback, as shown in Figures~\ref{prompt:eval_gen_sugg} and ~\ref{prompt:eval_human}.

\section{Data Leakage Analysis}
\label{appendix:data_leakage}

We use 4-grams and 5-grams to compute the similarity between the instructions in the training set and the questions in the benchmark. And then we draw scatter plots by selecting the top-200 data points with the highest similarity from each dataset, as shown in Figure~\ref{fig:dataleakage}. The scatter plot visually illustrates the data leakage phenomenon in the dataset, primarily concentrated in the HumanEval (+) and the latter half of the MBPP (+). This makes it difficult for our model to surpass baselines on these two benchmarks. In contrast, on the leakage-free LiveCodeBench and BigCodeBench-hard benchmarks, our model can outperform all baselines.



\section{Diminishing Returns}

\begin{figure}[]
    \centering
    \includegraphics[width=1\columnwidth]{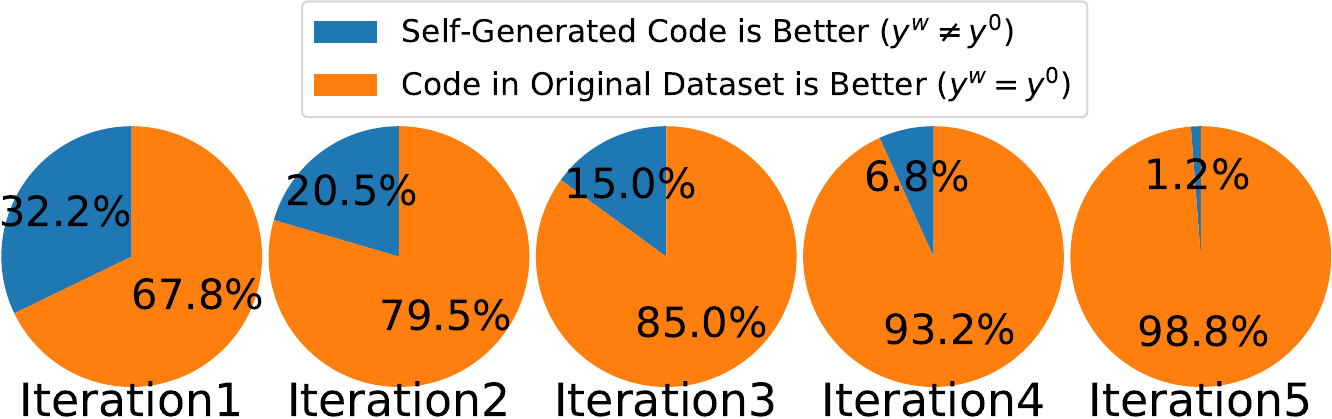}
    \caption{Proportions of two types of data in the iterative process.}
		\label{exp:pie_data_ratio}
\end{figure}

As shown in Figure~\ref{exp:pie_data_ratio}, we present the proportions of two types of data during the iterative process of RefineCoder-QW-7B. The self-generated code struggles more to outperform as iterations continue, indicating diminishing potential for further gains through self-refinement. The corresponding experimental results are shown in Figure 6. As observed, the model exhibits clear diminishing returns after the third iteration. This suggests that the model may have entered a mild overfitting stage, where further refinement not only fails to improve performance but may also exacerbate the issue.

\begin{figure}[t]
    \centering
    \includegraphics[width=1\linewidth]{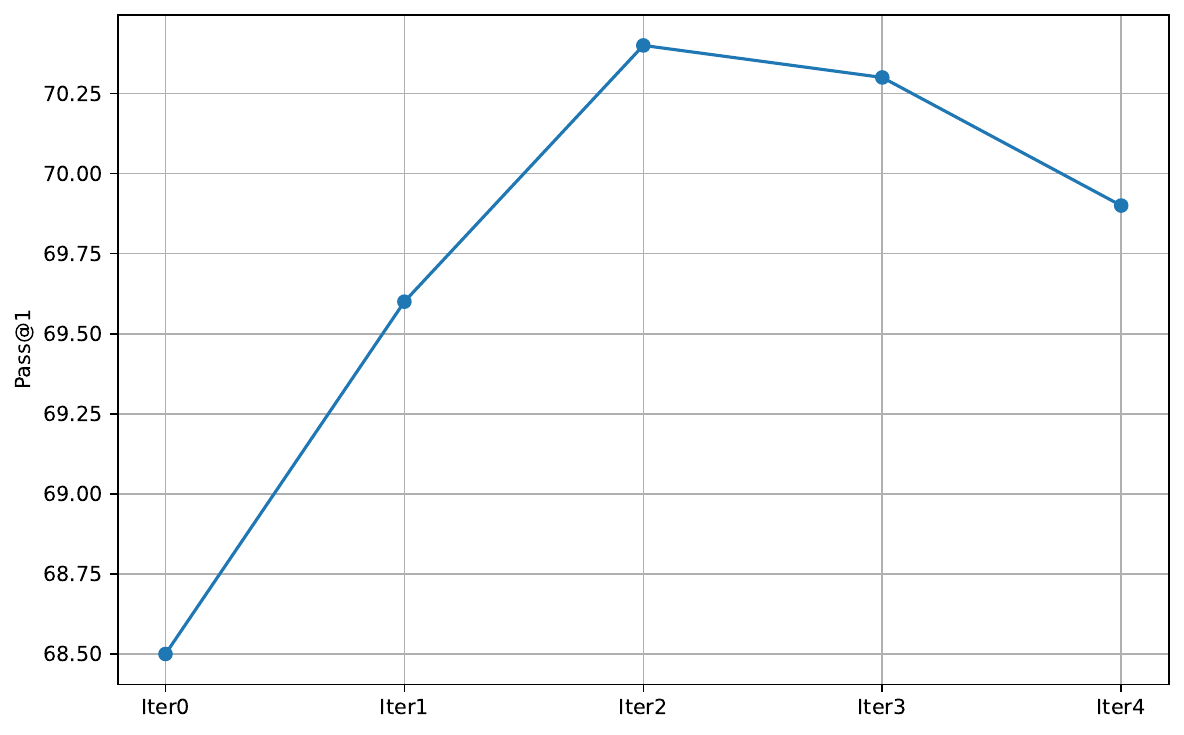}
    \caption{The average Pass@1 value of RefineCoder-QW-7B on HumanEval(+), MBPP(+) and BigCodeBench.}
    \label{exp:ablation_zhexian}
\end{figure}


\begin{figure*}[t]
    \centering
    \begin{minipage}{0.25\textwidth}
        \centering
        \includegraphics[width=\linewidth]{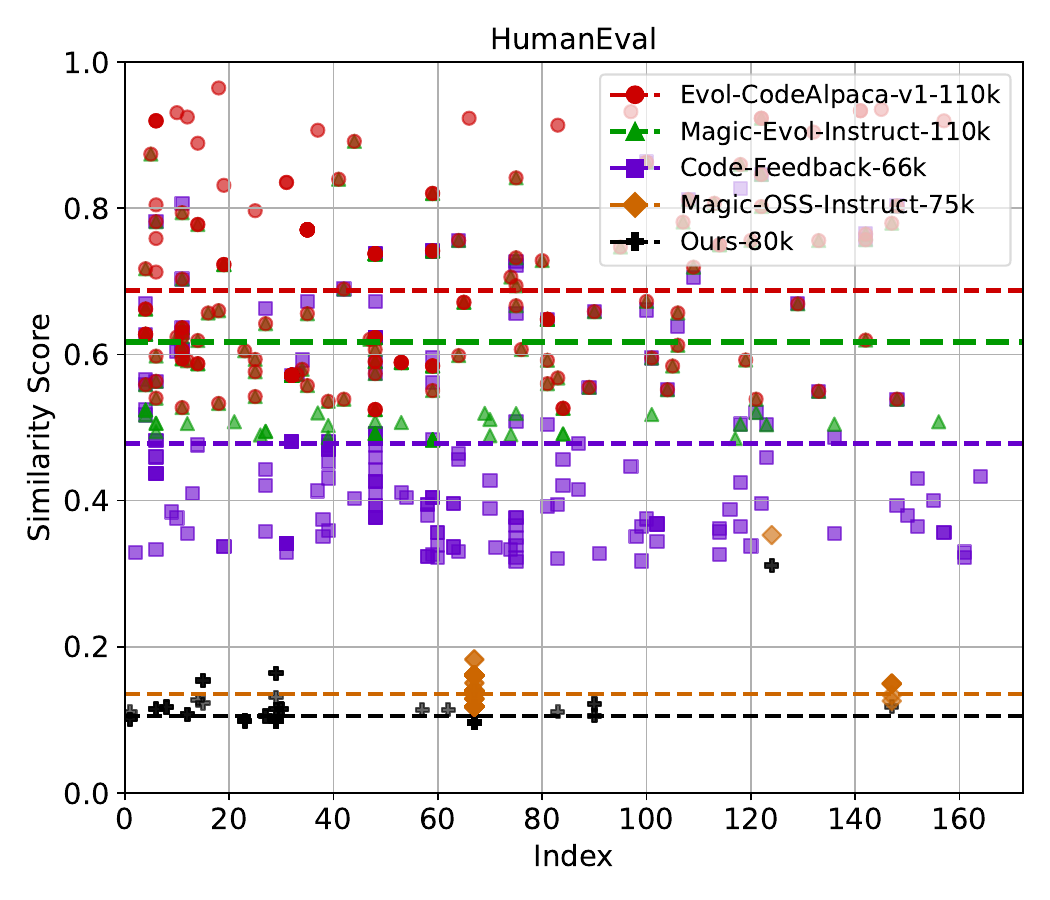}
    \end{minipage}%
    \begin{minipage}{0.25\textwidth}
        \centering
        \includegraphics[width=\linewidth]{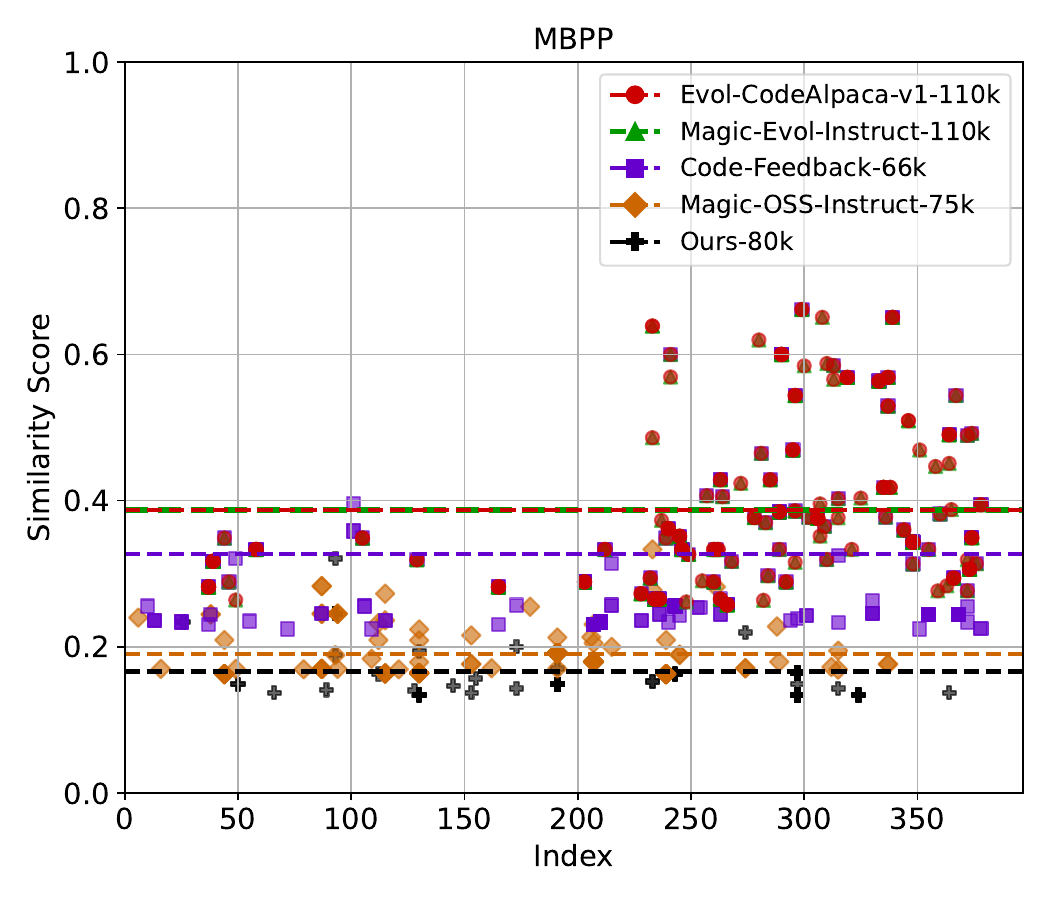}
    \end{minipage}%
    \begin{minipage}{0.25\textwidth}
        \centering
        \includegraphics[width=\linewidth]{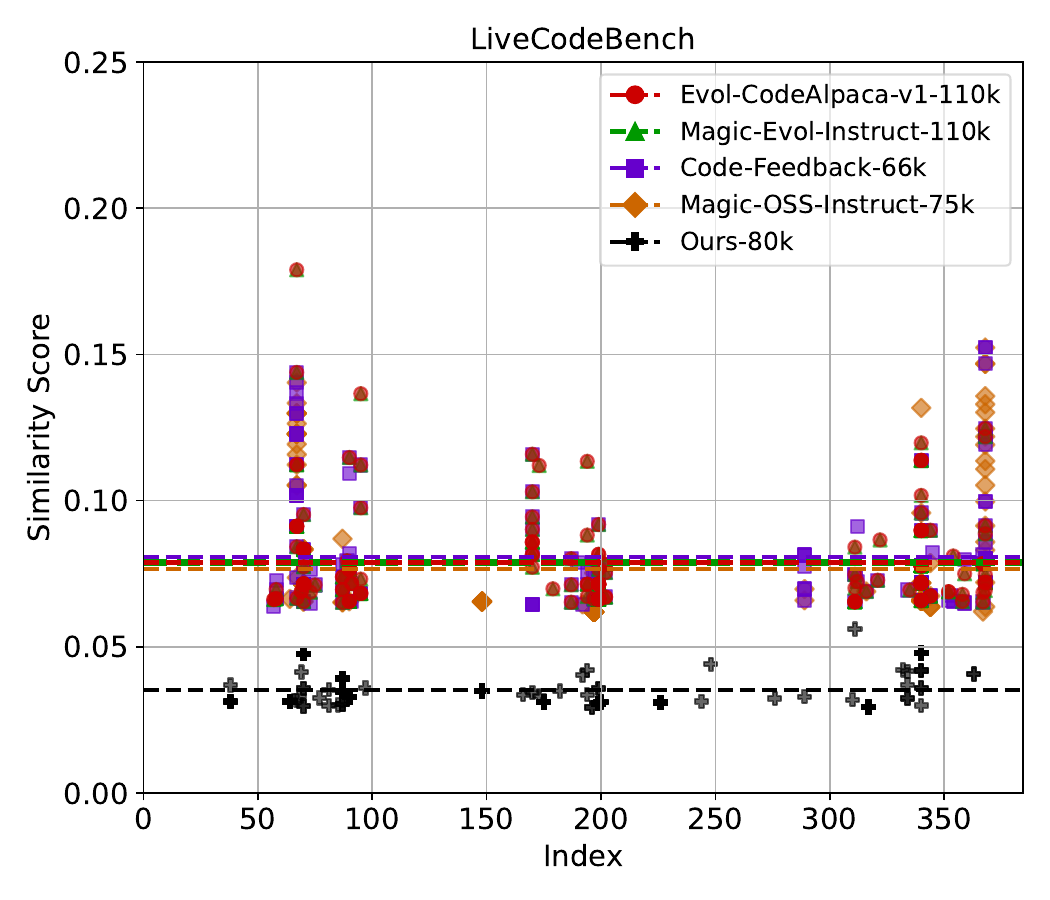}
    \end{minipage}%
    \begin{minipage}{0.25\textwidth}
        \centering
        \includegraphics[width=\linewidth]{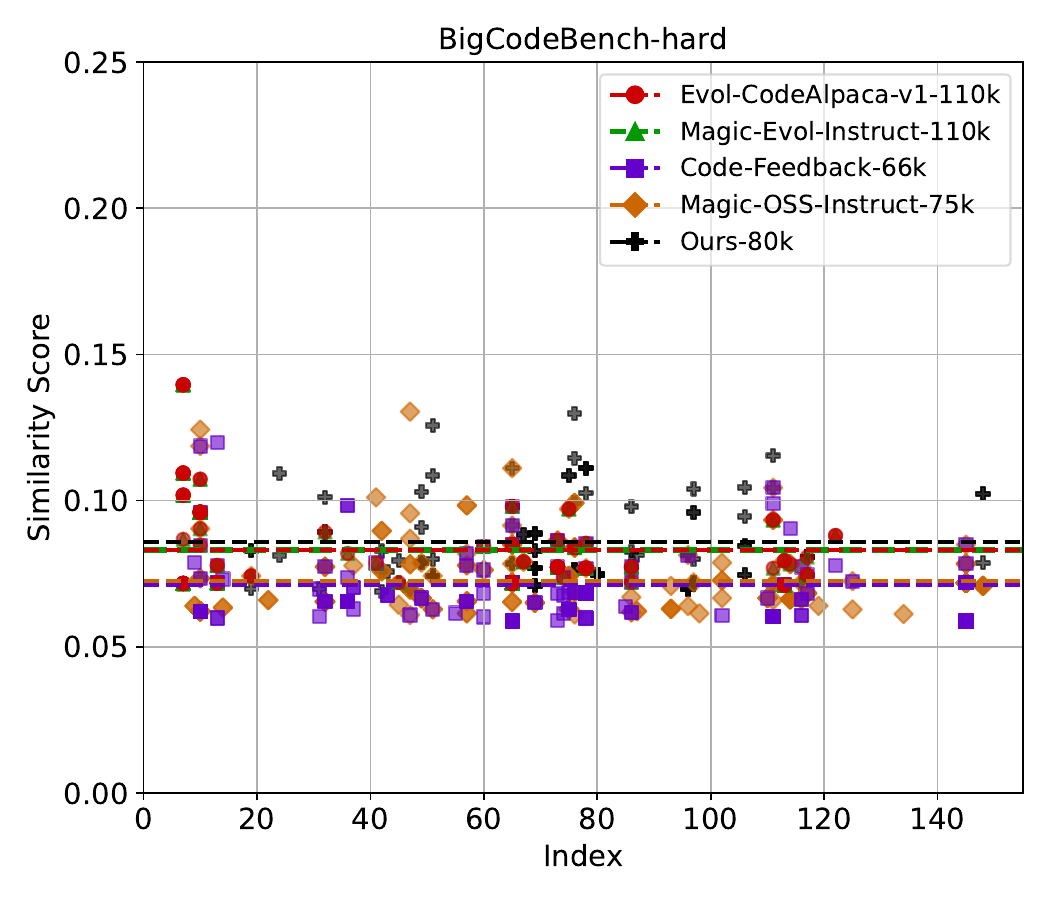}
    \end{minipage}
    \caption{Scatter plot of similarity scores between the datasets and four benchmarks, with average similarity scores for different datasets indicated by horizontal dashed lines. We selected the top 200 data points with the highest similarity from each dataset. Different y-axis ranges are set for better visualization.}
    \label{fig:dataleakage}
\end{figure*}


\begin{figure*}[]
    \centering
    \includegraphics[width=2.0\columnwidth]{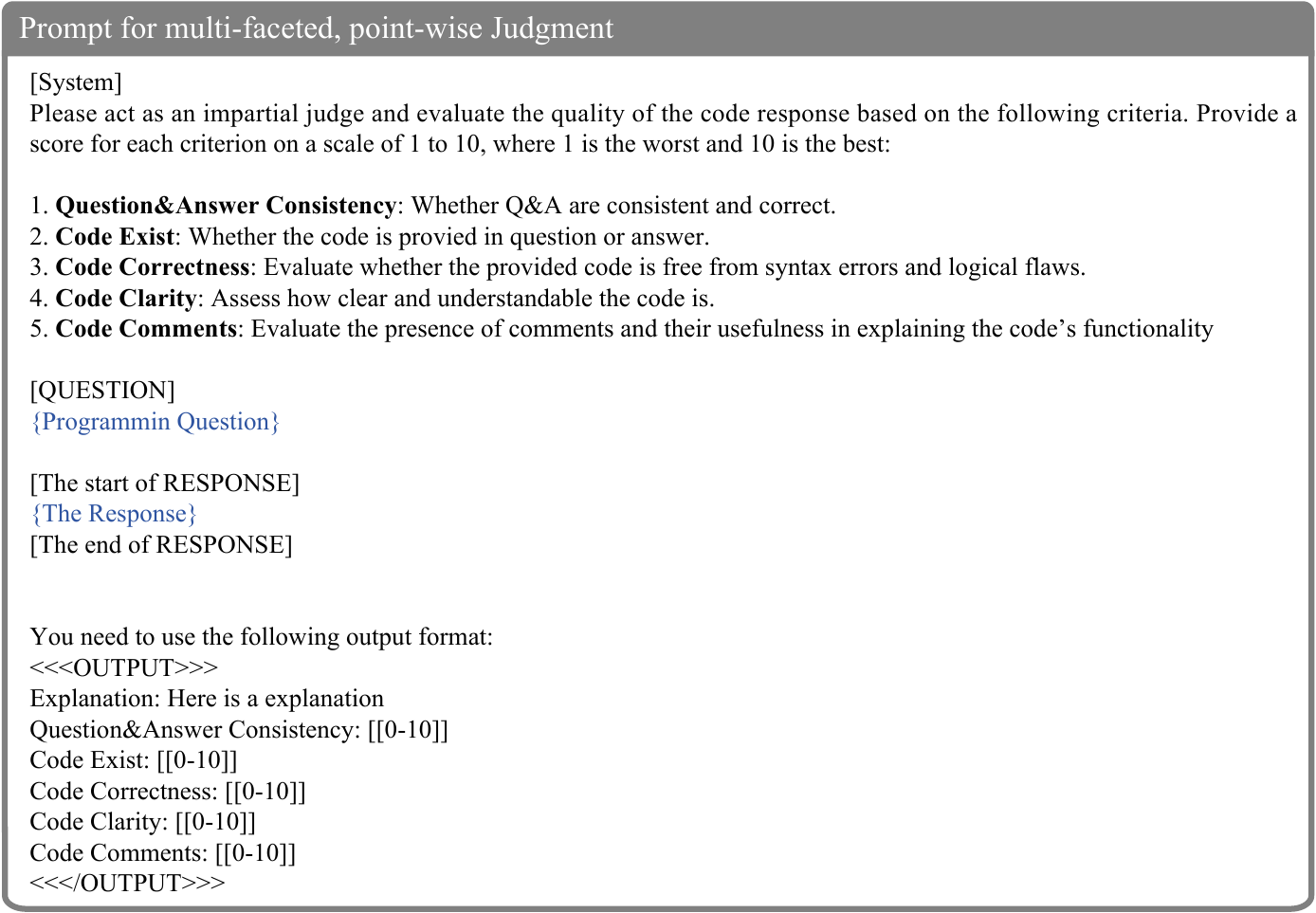}
    \caption{The prompt template for pointwise judgment.}
		\label{prompt:pointjudge}
\end{figure*}

\begin{figure*}[]
    \centering
    \includegraphics[width=2.0\columnwidth]{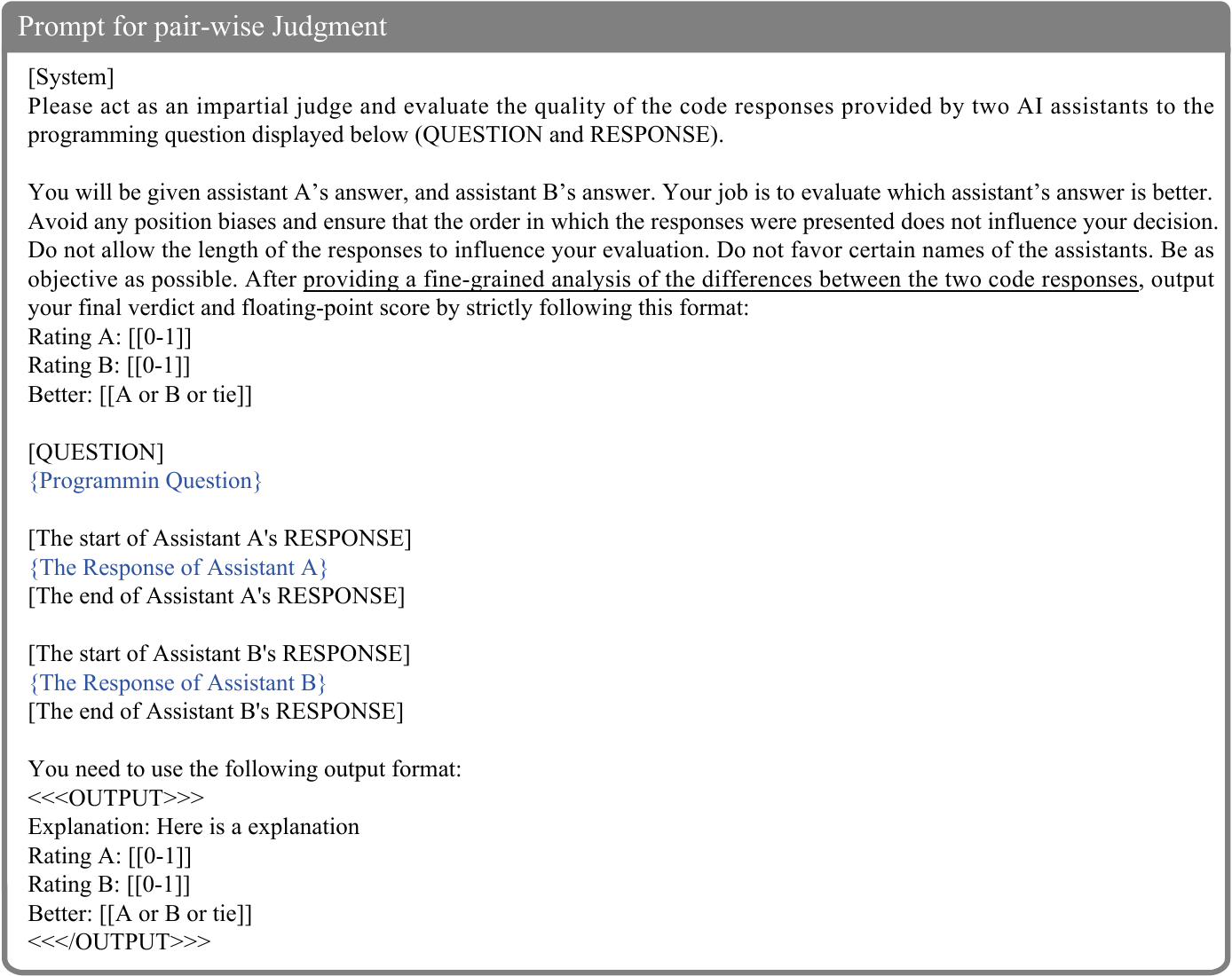}
    \caption{The prompt template for pairwise judgment.}
		\label{prompt:pairjudge}
\end{figure*}

\begin{figure*}[]
    \centering
    \includegraphics[width=2.0\columnwidth]{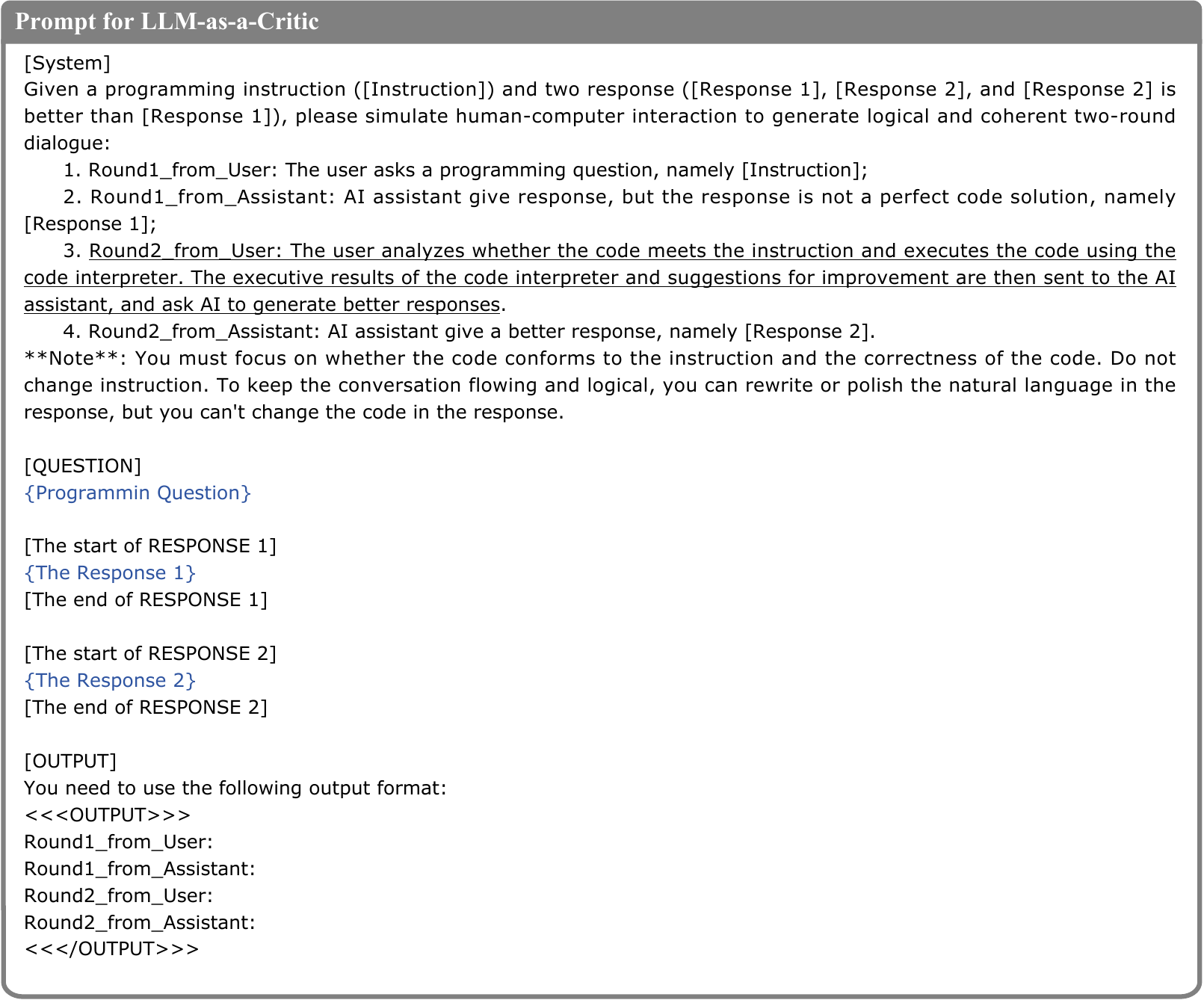}
    \caption{The prompt template for pairwise critique.}
		\label{prompt:paircritic}
\end{figure*}

\begin{figure*}[]
    \centering
    \includegraphics[width=2.0\columnwidth]{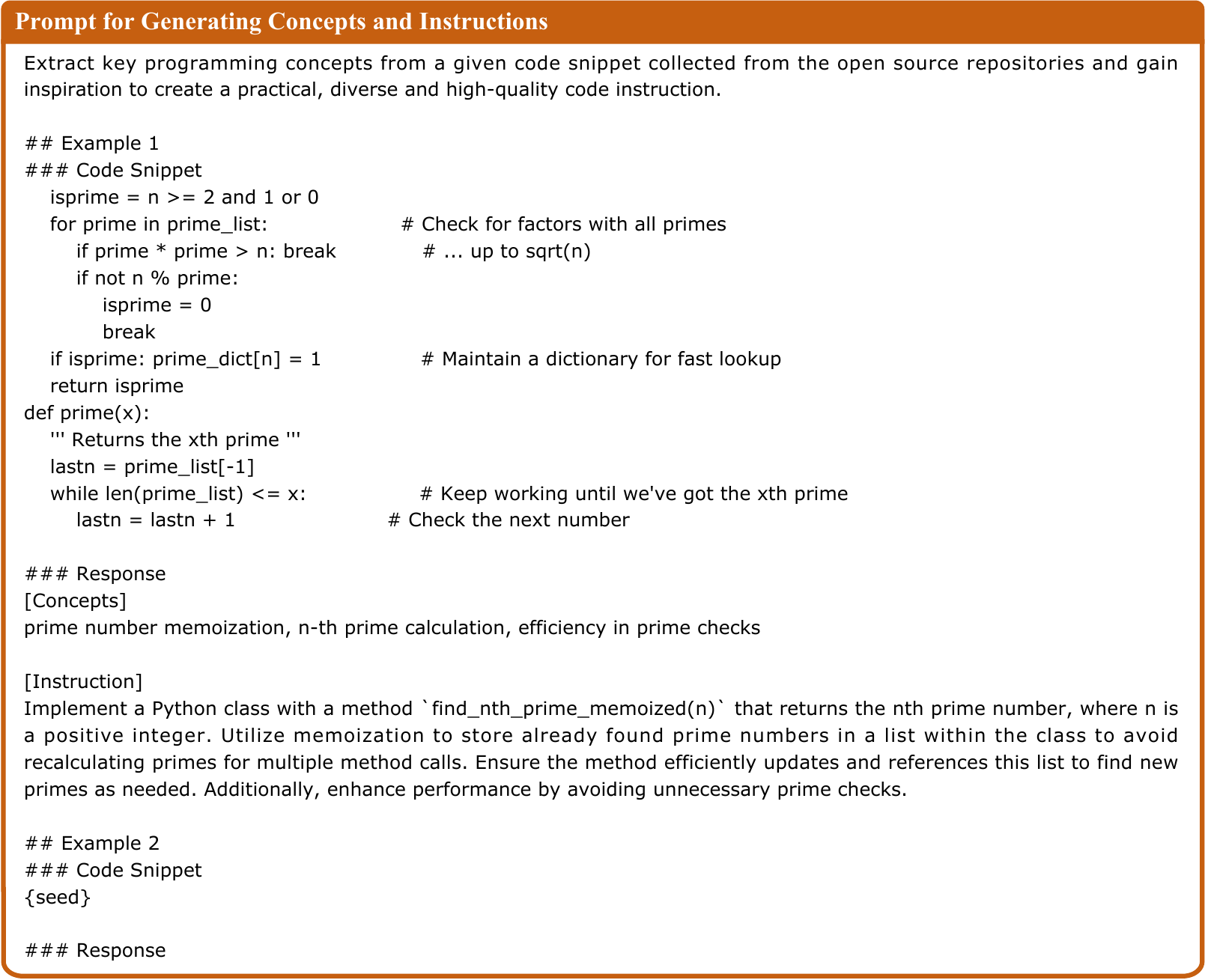}
    \caption{The prompt template for generating code-related concepts and instructions.}
    \label{prompt:gen_ins}
\end{figure*}

\begin{figure*}[]
    \centering
    \includegraphics[width=2.0\columnwidth]{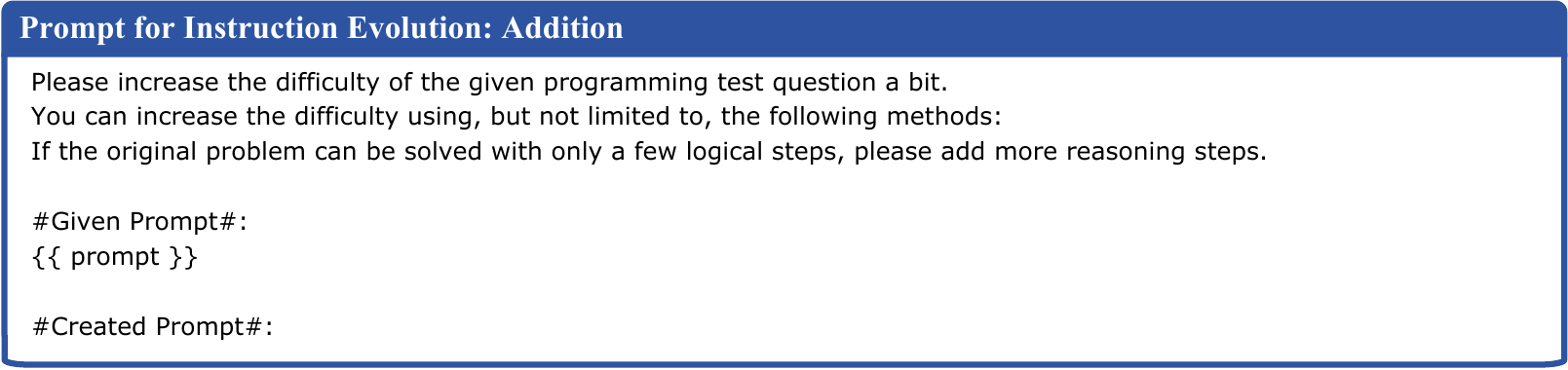}
    \caption{The prompt template for Addition evolution.}
    \label{prompt:evol_addition}
\end{figure*}

\begin{figure*}[]
    \centering
    \includegraphics[width=2.0\columnwidth]{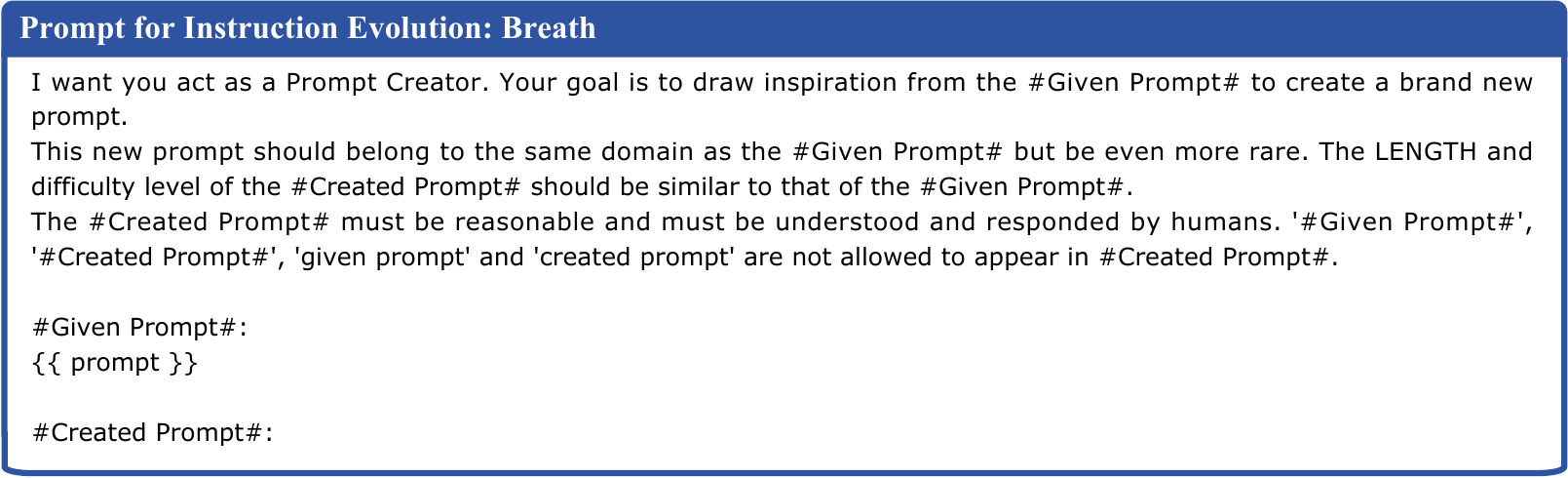}
    \caption{The prompt template for Breath evolution.}
    \label{prompt:evol_breath}
\end{figure*}

\begin{figure*}[]
    \centering
    \includegraphics[width=2.0\columnwidth]{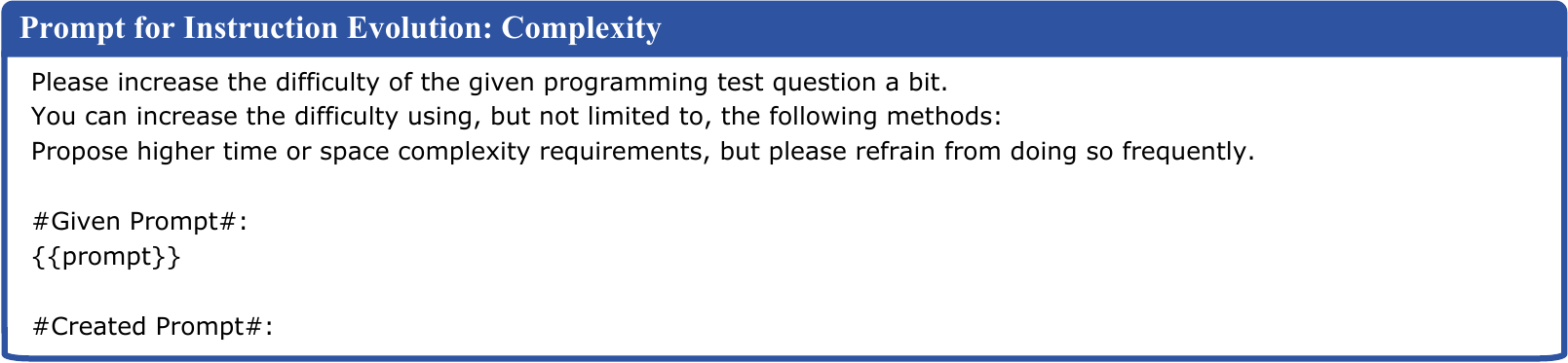}
    \caption{The prompt template for Complexity evolution.}
    \label{prompt:evol_comp}
\end{figure*}
\begin{figure*}[]
    \centering
    \includegraphics[width=2.0\columnwidth]{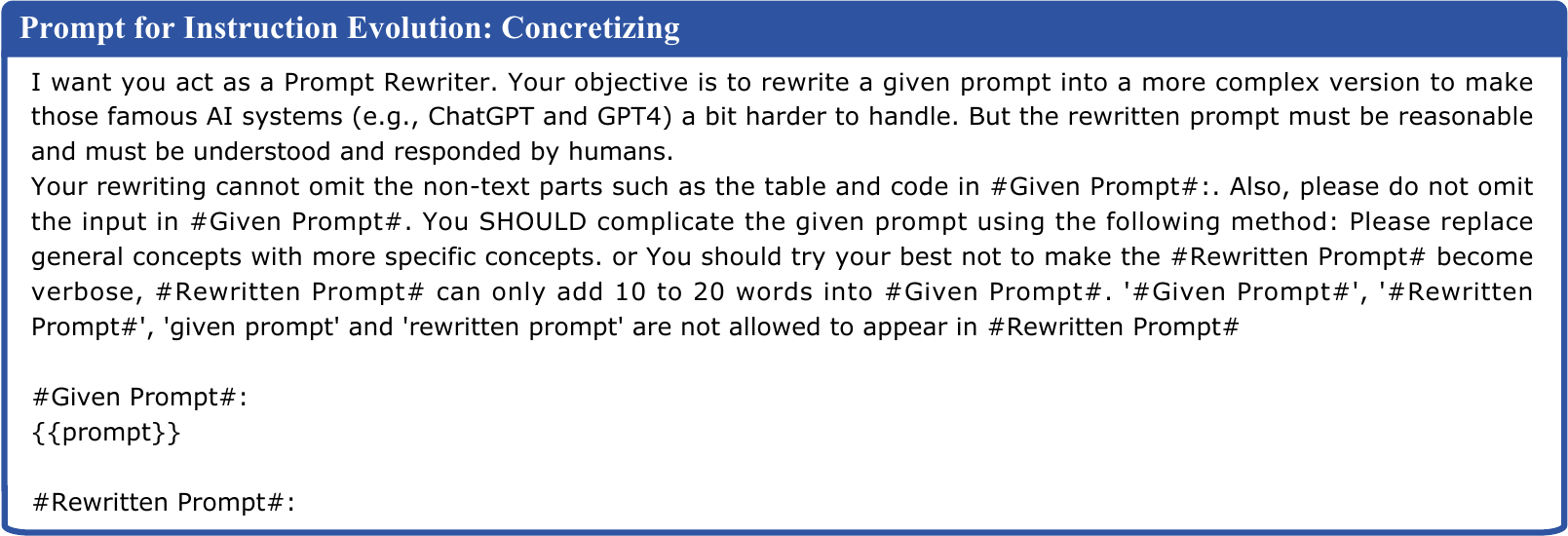}
    \caption{The prompt template for Concretizing evolution.}
    \label{prompt:evol_concre}
\end{figure*}
\begin{figure*}[]
    \centering
    \includegraphics[width=2.0\columnwidth]{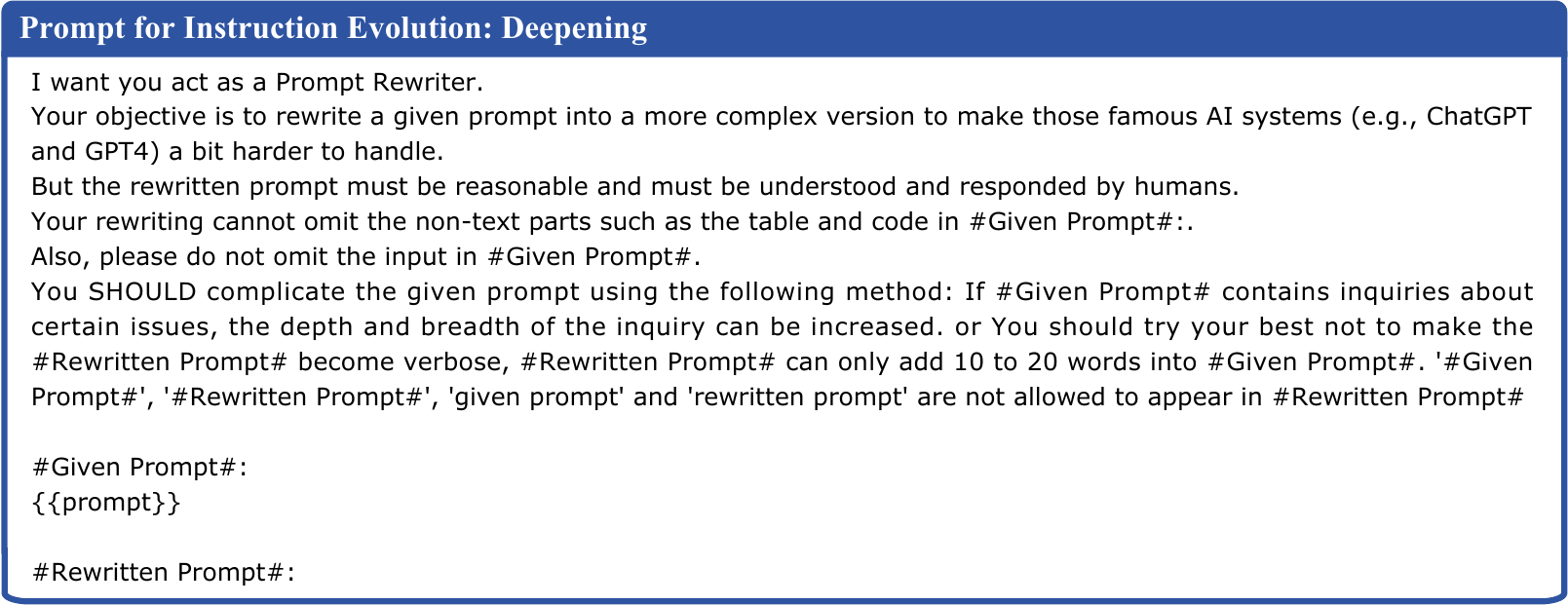}
    \caption{The prompt template for Deepening evolution.}
    \label{prompt:evol_deep}
\end{figure*}

\begin{figure*}[]
    \centering
    \includegraphics[width=2.0\columnwidth]{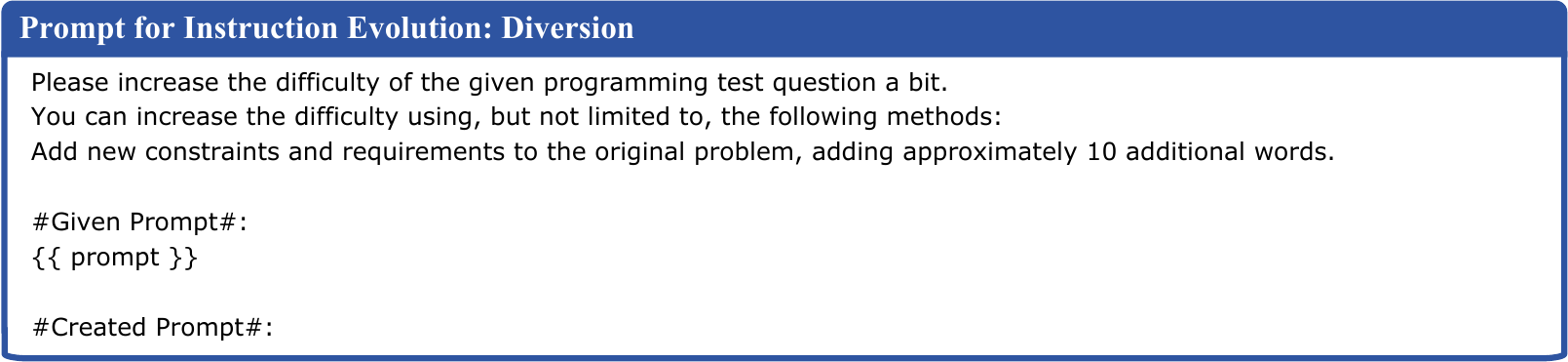}
    \caption{The prompt template for Diversion evolution.}
    \label{prompt:evol_diversion}
\end{figure*}

\begin{figure*}[]
    \centering
    \includegraphics[width=2.0\columnwidth]{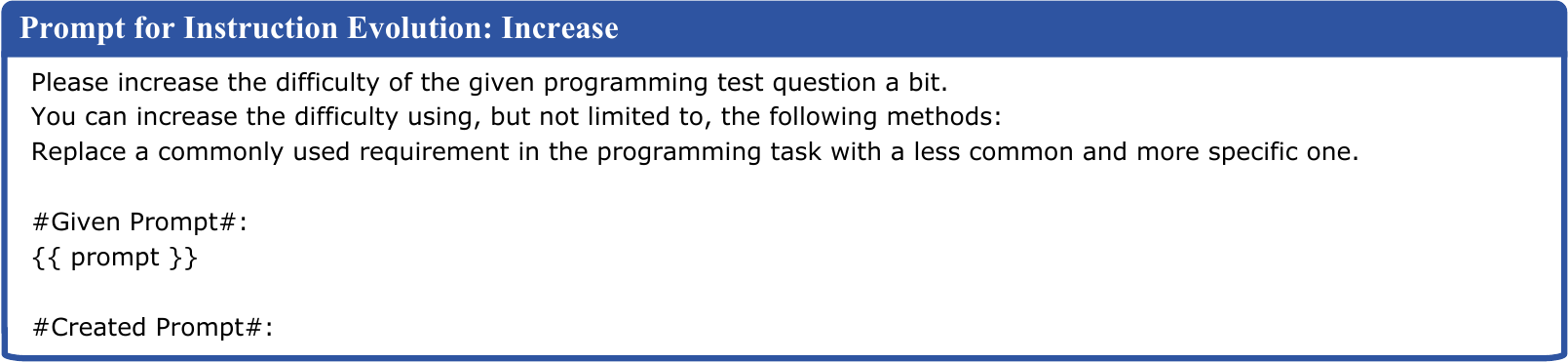}
    \caption{The prompt template for Increase evolution.}
    \label{prompt:evol_increase}
\end{figure*}

\begin{figure*}[]
    \centering
    \includegraphics[width=2.0\columnwidth]{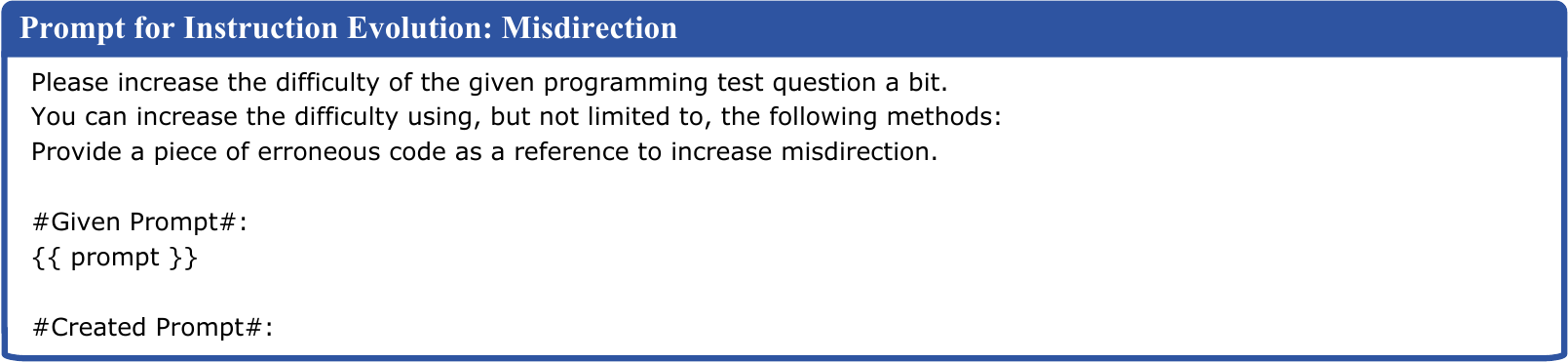}
    \caption{The prompt template for Misdirection evolution.}
    \label{prompt:evol_misd}
\end{figure*}

\begin{figure*}[]
    \centering
    \includegraphics[width=2.0\columnwidth]{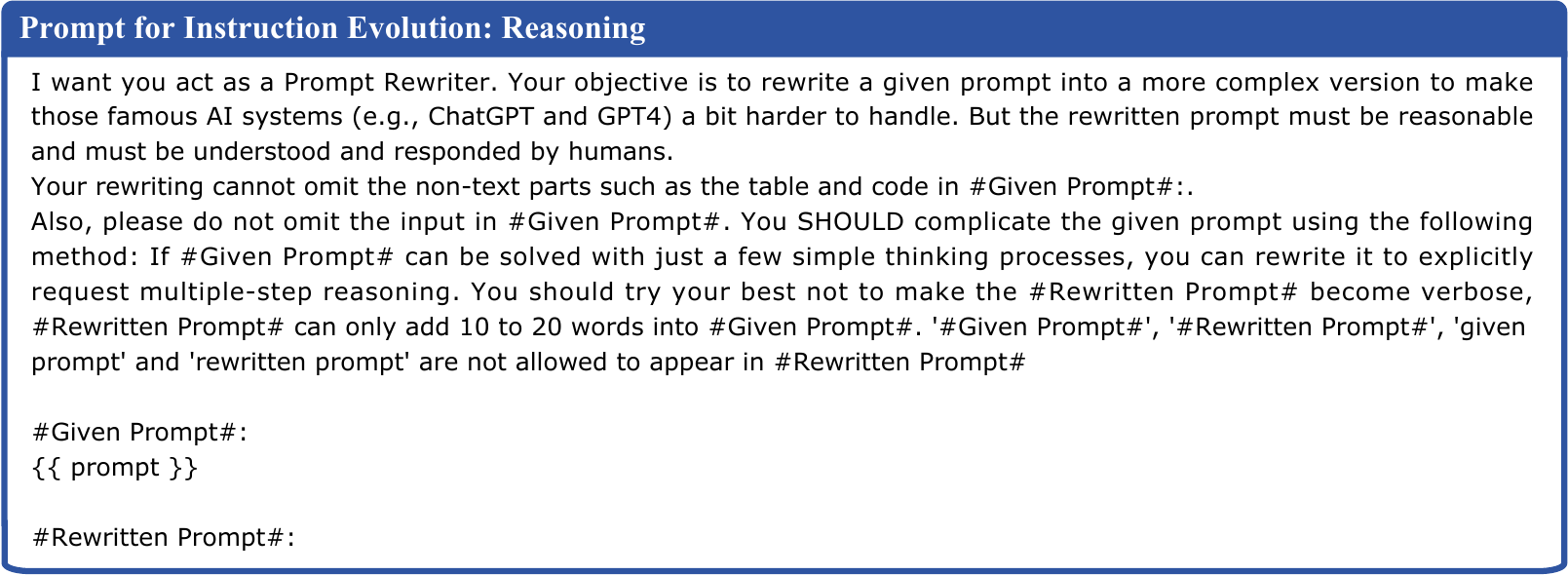}
    \caption{The prompt template for Reason evolution.}
    \label{prompt:evol_reason}
\end{figure*}

\begin{figure*}[]
    \centering
    \includegraphics[width=2.0\columnwidth]{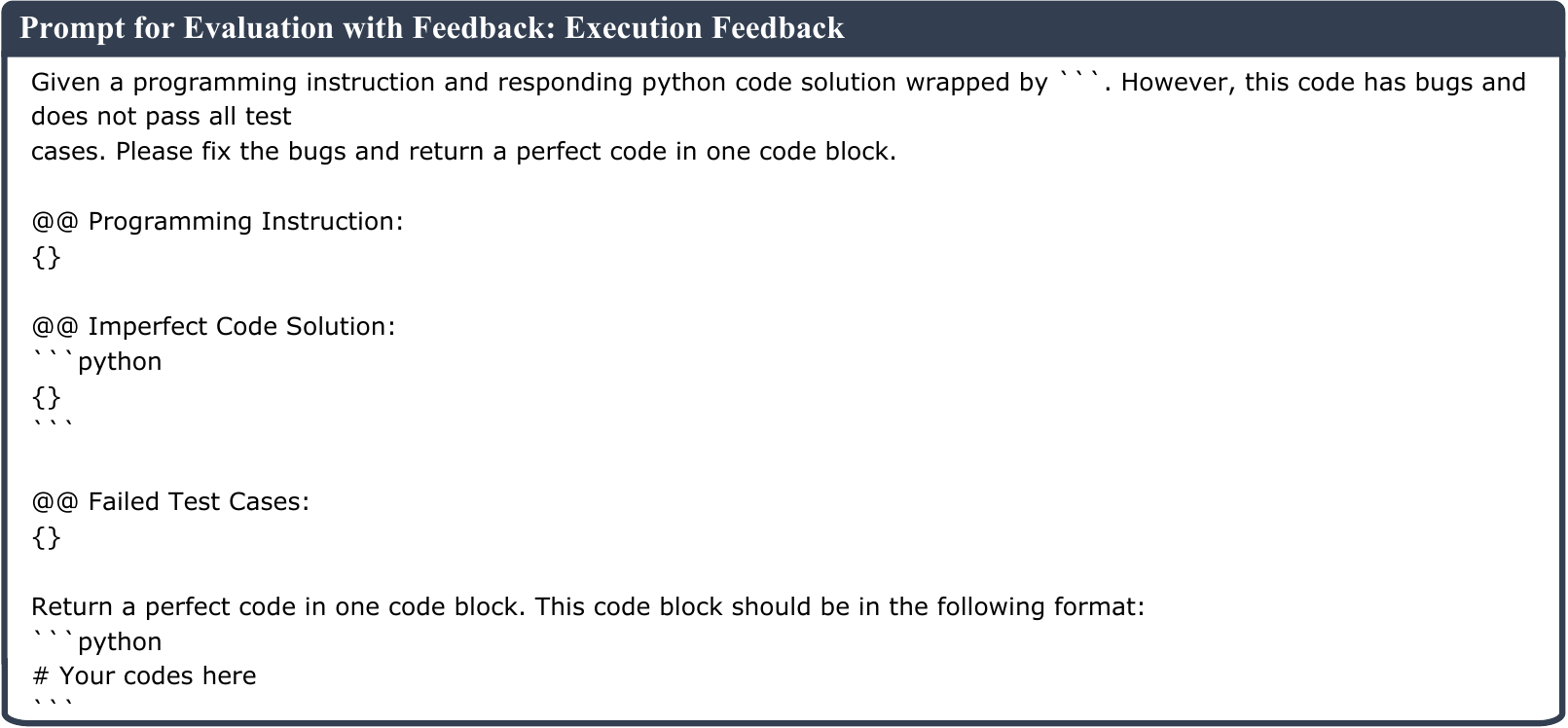}
    \caption{The prompt template for Execution Feedback.}
    \label{prompt:eval_exec}
\end{figure*}

\begin{figure*}[]
    \centering
    \includegraphics[width=2.0\columnwidth]{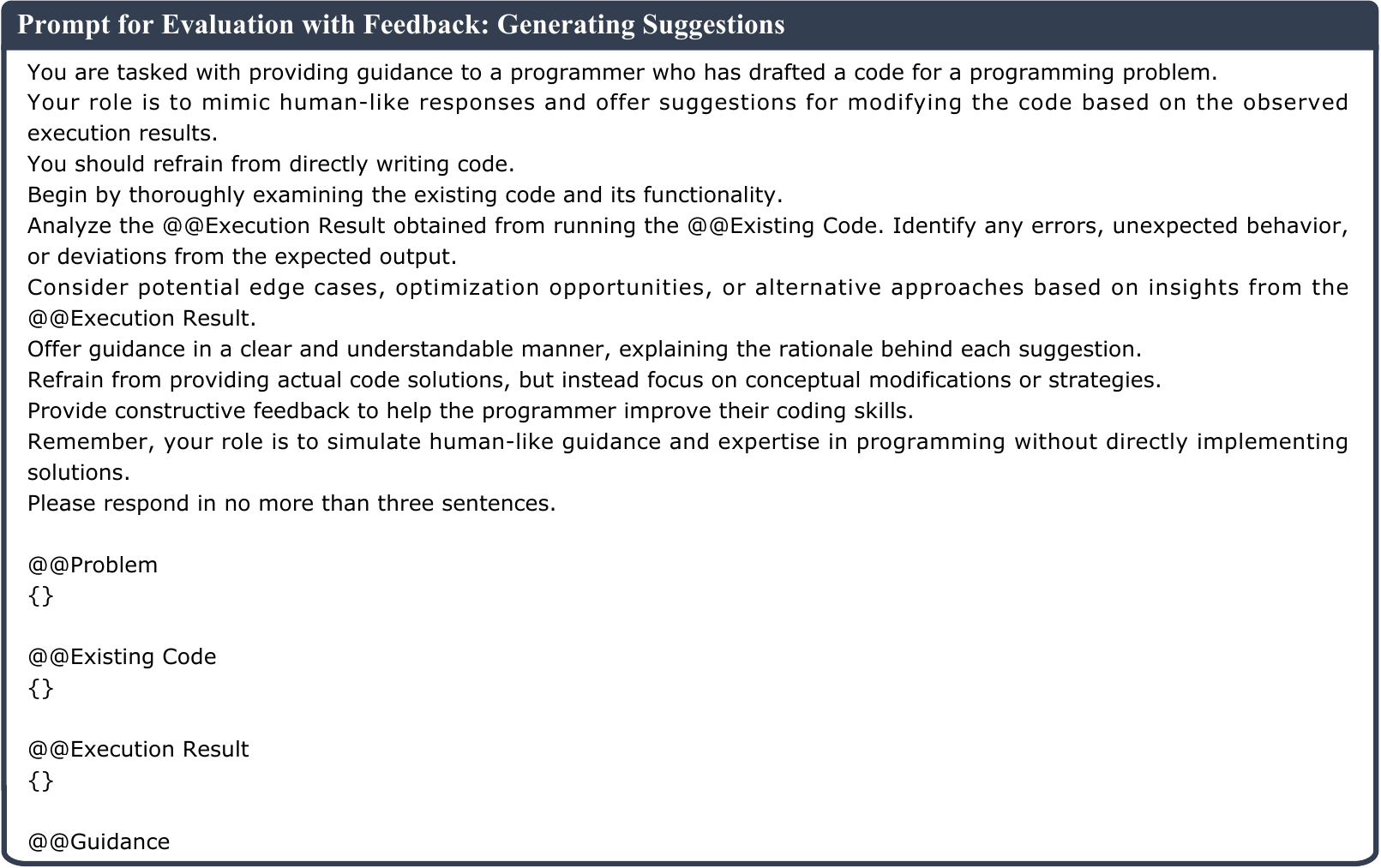}
    \caption{The prompt template for generating improvement suggestions.}
    \label{prompt:eval_gen_sugg}
\end{figure*}

\begin{figure*}[]
    \centering
    \includegraphics[width=2.0\columnwidth]{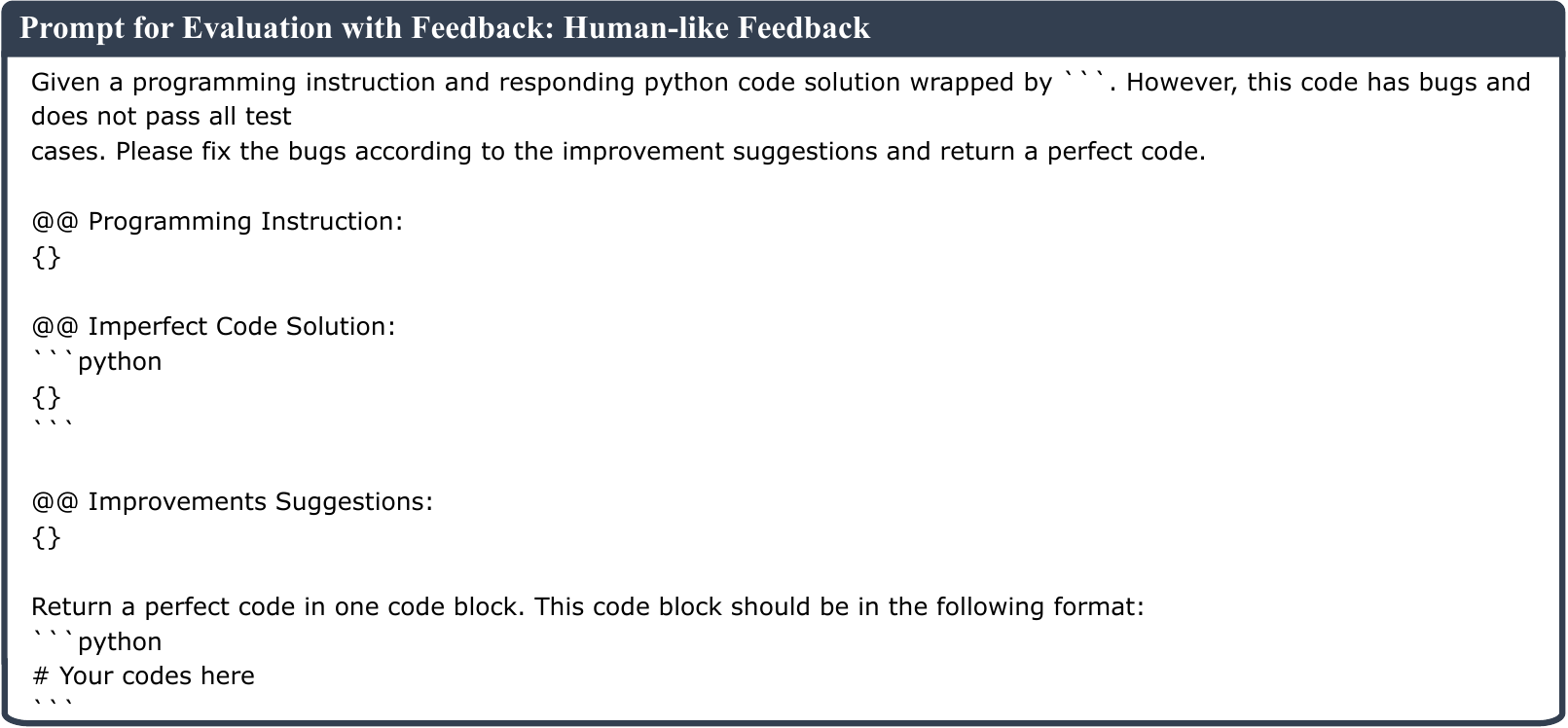}
    \caption{The prompt template for Human-like Feedback.}
    \label{prompt:eval_human}
\end{figure*}

\end{document}